\documentclass[sigconf]{acmart}
\usepackage[noend]{algpseudocode}
\usepackage{algorithmicx,algorithm}
\usepackage{float}
\usepackage{balance}

\AtBeginDocument{%
  \providecommand\BibTeX{{%
    \normalfont B\kern-0.5em{\scshape i\kern-0.25em b}\kern-0.8em\TeX}}}

\setcopyright{acmcopyright}
\copyrightyear{2023} 
\acmYear{2023} 
\setcopyright{acmlicensed}\acmConference[MM '23]{Proceedings of the 31st ACM International Conference on Multimedia}{October 29-November 3, 2023}{Ottawa, ON, Canada}
\acmBooktitle{Proceedings of the 31st ACM International Conference on Multimedia (MM '23), October 29-November 3, 2023, Ottawa, ON, Canada}
\acmPrice{15.00}
\acmDOI{10.1145/3581783.3612022}
\acmISBN{979-8-4007-0108-5/23/10}

\acmSubmissionID{1445}
\begin{document}

\title{AvatarFusion: Zero-shot Generation of Clothing-Decoupled 3D Avatars Using 2D Diffusion}

\author{Shuo Huang}
\email{huangs22@mails.tsinghua.edu.cn}
\affiliation{%
  \institution{Department of Computer Science and
Technology, Tsinghua University}
  \city{Beijing}
  \country{China}
}

\author{Zongxin Yang}
\email{yangzongxin@zju.edu.cn}
\affiliation{%
  \institution{ReLER, CCAI, Zhejiang University}
  \city{Zhejiang}
  \country{China}
}

\author{Liangting Li}
\email{llt21@mails.tsinghua.edu.cn}
\affiliation{%
  \institution{Department of Computer Science and
Technology, Tsinghua University}
  \city{Beijing}
  \country{China}
}

\author{Yi Yang}
\email{yangyics@zju.edu.cn}
\affiliation{%
 \institution{ReLER, CCAI, Zhejiang University}
 \city{Zhejiang}
  \country{China}
}

\author{Jia Jia}
\authornote{Corresponding author.}
\email{jjia@tsinghua.edu.cn}
\affiliation{%
  \institution{Department of Computer Science and
Technology, Tsinghua University}
  \institution{Beijing National Research Center for Information Science and Technology}
  \city{Beijing}
  \country{China}
}

\renewcommand{\shortauthors}{Shuo Huang et al.}
\begin{abstract}
Large-scale pre-trained vision-language models allow for the zero-shot text-based generation of 3D avatars. The previous state-of-the-art method utilized CLIP to supervise neural implicit models that reconstructed a human body mesh. However, this approach has two limitations. Firstly, the lack of avatar-specific models can cause facial distortion and unrealistic clothing in the generated avatars. Secondly, CLIP only provides optimization direction for the overall appearance, resulting in less impressive results. To address these limitations, we propose AvatarFusion, the first framework to use a latent diffusion model to provide pixel-level guidance for generating human-realistic avatars while simultaneously segmenting clothing from the avatar's body. AvatarFusion includes the first clothing-decoupled neural implicit avatar model that employs a novel Dual Volume Rendering strategy to render the decoupled skin and clothing sub-models in one space. We also introduce a novel optimization method, called Pixel-Semantics Difference-Sampling (PS-DS), which semantically separates the generation of body and clothes, and generates a variety of clothing styles. Moreover, we establish the first benchmark for zero-shot text-to-avatar generation. Our experimental results demonstrate that our framework outperforms previous approaches, with significant improvements observed in all metrics. Additionally, since our model is clothing-decoupled, we can exchange the clothes of avatars. Code are available on our project page \href{https://hansenhuang0823.github.io/AvatarFusion}{https://hansenhuang0823.github.io/AvatarFusion}.
\end{abstract}



\begin{teaserfigure}
\centering
    \includegraphics[width=\textwidth]{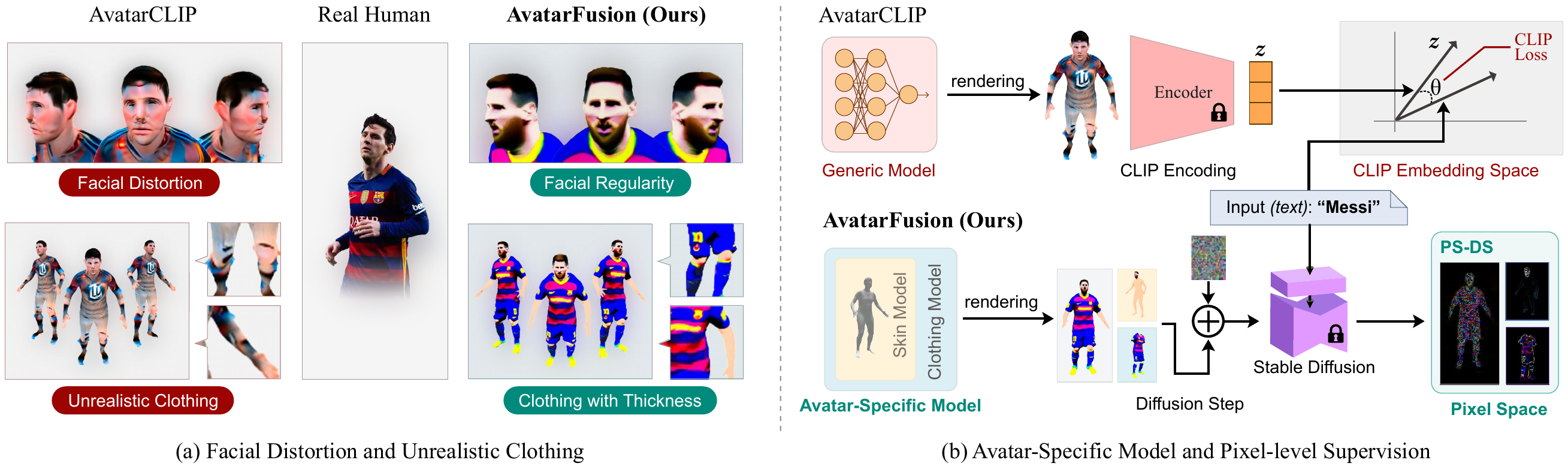}
    \caption{Given a text prompt "Messi", our method (AvatarFusion) effectively alleviates (a) the problem of facial distortion and unrealistic clothing and generates more photo-realistic avatars. To achieve this, we introduce (b) an avatar-specific model and a pixel-level diffusion supervision (PS-DS) which separates the generation of skin and clothing for better realism.}
    \label{fig:problem}
\end{teaserfigure}
\maketitle

\section{Introduction}

Recently, the zero-shot text-to-avatar generation task has become feasible due to the emergence of large-scale vision-language models \cite{radford2021learning, ramesh2022hierarchical, saharia2022photorealistic, rombach2022high}. This task utilizes these pre-trained models as supervision to create highly compelling avatars that represent celebrities or virtual novel characters by simply inputting
text. Compared with the 3D GANs currently evolving for human body generation \cite{chan2022efficient, zhang2023avatargen, or2022stylesdf}, zero-shot text-to-avatar generation does not require large amounts of richly-annotated human body datasets, extensive computing resources, or the training of difficult-to-converge 3D human body generations. Furthermore, it can generate more diverse avatars, making it a promising technology with significant potential for various applications. However, effectively integrating generalizable large-scale models with parametric human models to leverage multi-knowledge representations~\cite{mkr} remains an open problem for improving avatar generation.

The previous methods of generating avatars from text were limited to 2D representations \cite{jiang2022text2human, wang2018high, park2019semantic, yoon2021pose, lewis2021tryongan, weng2020misc}. This was due to the lack of 3D datasets. However, recent advancements in differentiable rendering \cite{mildenhall2021nerf, li2018differentiable, muller2019neural, niemeyer2020differentiable} have made it possible to supervise a 3D model by using a vision-language model to supervise its rendered images. This has led to the development of zero-shot text-to-3D object methods \cite{poole2022dreamfusion, michel2022text2mesh, lin2022magic3d, wang2022rodin, wang2022nerf, nichol2022point, jain2022zero}.
Despite the progress in generating common 3D objects, generating 3D avatars still poses significant challenges. The flexible movements and complex body structure of 3D avatars make it difficult for these models to generate avatars that exhibit optimal performance. However, AvatarCLIP \cite{hong2022avatarclip} was specifically designed to address these issues. It uses an implicit neural network to reconstruct a human body prior \cite{loper2015smpl} and leverages CLIP (Contrastive Language-Image Pretraining) \cite{radford2021learning}, a vision-language model, to optimize the reconstructed human body towards the text description. In this approach, CLIP encodes both the text and the rendered image to embeddings in a multi-modal feature space, and then minimizes their cosine distance.

Despite significant advancements made by AvatarCLIP in the zero-shot text-to-avatar task, there are still several limitations that require attention.
1) A lack of avatar-specific model. Current methods do not employ specialized models designed for avatars. Instead, they rely on generic 3D models, which leads to two major issues: facial distortion and unrealistic clothing, which greatly compromise the quality of the generated avatars as shown in Figure \ref{fig:problem}. 
Facial distortion arises from a poor combination of these models with human body priors, resulting in imprecise human body representations with shapeless faces. Unrealistic clothing is caused by the current models treating clothing and skin as mere textures on the body surface, without differentiating between clothing and body models, which results in a mixed texture and a lack of clothing thickness.
2) Limited generative power of CLIP. Although CLIP has successfully bridged the gap between natural language and avatar images, it has limited generative power as it is not a generative model. It only calculates an embedding of the overall appearance, which causes generated avatars to mismatch with the text in terms of details.

To address the first limitation, it is necessary to develop more sophisticated avatar models that can also effectively separate clothing and skin textures to simulate realistic clothing. One promising approach is the use of clothing-decoupled models, such as those proposed in \cite{feng2022capturing, corona2021smplicit, jiang2020bcnet, zhu2020deep, xiang2021modeling}. However, these models do not use a complete neural implicit representation, which has been shown to be the most suitable 3D representation for receiving guidance from vision-language models \cite{michel2022text2mesh, hong2022avatarclip}.
As for the second limitation, one possible solution is to use diffusion models' supervision \cite{poole2022dreamfusion, lin2022magic3d} to optimize the human body model, as diffusion models are pixel-level generative models. However, due to the lack of ground truth annotated with clothing segmentation labels, existing clothing-decoupled models cannot obtain an optimization direction that effectively distinguishes clothing from skin through current optimization methods.

To tackle these challenges, we present AvatarFusion, a pioneering method for generating clothing-decoupled 3D avatars from text prompts using diffusion models as supervision. Notably, to the best of our knowledge, our approach is the first to combine zero-shot 3D avatar generation and segmentation.
In AvatarFusion, we present the first clothing-decoupled avatar model of complete neural implicit representations, along with a novel Dual Volume Rendering strategy for its rendering. Additionally, we propose a novel optimization loss called Pixel-Semantics Difference-Sampling (PS-DS) to optimize the model from diffusion models and separate the generation of clothing and skin.
To elaborate on our proposed approach, we first utilize an off-the-shelf Signed Distance Function (SDF) field to create a SDF-based Avatar Model (SAvM) that accurately captures the intricate details of the human body prior, SMPL \cite{loper2015smpl}. SAvM is also equipped with a deformation field that allows for the manipulation of the avatar's pose during training.
Secondly, we introduce a clothing-decoupled model (CDM) that uses two SAvMs to represent the skin and clothing separately. Our proposed Dual Volume Rendering strategy is compatible with traditional volume rendering and provides a way to jointly render two implicit neural representations in the same space.
Finally, our proposed Pixel-Semantics Difference-Sampling (PS-DS) aligns the \textbf{D}ifferences in \textbf{P}ixel space to the \textbf{D}ifference of text \textbf{S}emantics as a \textbf{S}ampling strategy of a latent diffusion model known as Stable Diffusion \cite{rombach2022high}. This allows the clothing model to generate only the clothing around the avatar's body without covering the skin.

Prior to our work, evaluation metrics for this field were limited to user studies, and no benchmark was available. To fill this gap, we propose a new benchmark called Famous-Character-50 (FC50). It comprises fifty varied text prompts featuring famous people or fictional characters of different cultures and genders. Our benchmark enables quantitative evaluation by comparing the generated avatars with the outcomes of mature 2D text-to-image models. To assess the generated avatar images' quality, we employ face recognition distance \cite{geitgey2017face} and Fréchet Inception Distance (FID) score \cite{Seitzer2020FID} as evaluation metrics.
Our experimental results indicate that AvatarFusion outperforms baselines in both quantitative and qualitative evaluations on our benchmark. Specifically, AvatarFusion attains a lower face recognition distance of 14.04\%, implying that the generated avatars are more faithful to the given text prompts. Furthermore, our method significantly enhances the FID score, indicating that AvatarFusion generates more realistic avatars. The model's high performance in user studies also suggests that the generated avatars align better with the human perspective.
Additionally, our avatar-specific model is clothing-decoupled, enabling us to exchange clothing between different characters.

Our contributions are as follows.
\begin{itemize}
    \item 
    We propose AvatarFusion, a novel framework for zero-shot 3D avatar generation that is the first to combine zero-shot clothing segmentation and avatar generation. AvatarFusion includes a clothing-decoupled neural implicit model, a Dual Volume Rendering strategy, and a PS-DS optimization method.
    
    \item
    We propose the use of diffusion-based optimization methods to enhance the visual details of the generated avatars, specifically by improving the distinction between clothing and skin.

    \item
    We also propose the first benchmark for the field, called Famous-Character-50 (FC50). Our experiments demonstrate the effectiveness of our proposed approach in comparison to state-of-the-art methods.
    
\end{itemize}

\section{Related Works}
\textbf{Large-scale vision-language models.}
Recently, CLIP \cite{radford2021learning} has significantly advanced in bridging the gap between natural language and images by providing a multi-modal embedding space for various downstream tasks such as image generation \cite{ramesh2022hierarchical}, segmentation \cite{xu2021simple, ding2022decoupling, wang2022cris,samtrack}, classification \cite{radford2021learning}, and captioning \cite{tewel2021zero, mokady2021clipcap}.
Text-conditioned diffusion models \cite{nichol2021glide, saharia2022photorealistic, ramesh2022hierarchical, rombach2022high} are another breakthrough. DALL-E 2 \cite{ramesh2022hierarchical} uses CLIP's embedding space to generate images of complex text prompts, while Imagen \cite{saharia2022photorealistic} employs a cascade of super-resolution models to improve generation efficiency. Stable Diffusion \cite{rombach2022high} takes a different approach by using a low-resolution latent space to generate images.

\noindent\textbf{Zero-shot text-to-3D-objects generation.}
Thanks to the recent development of pre-trained vision-language models, several works have contributed to generating 3D objects in a zero-shot manner from text prompts. Text2mesh \cite{michel2022text2mesh}, AvatarCLIP \cite{hong2022avatarclip}, and NeRF-Art \cite{wang2022nerf} utilize CLIP \cite{radford2021learning} loss as supervision to optimize mesh and implicit function representations.
In addition to CLIP, diffusion models \cite{ramesh2022hierarchical, saharia2022photorealistic, rombach2022high, zhou20213d} can also be used for text-based model supervision \cite{wang2022rodin, poole2022dreamfusion, lin2022magic3d, metzer2022latent, luo2021diffusion, nichol2022point, richardson2023texture}. DreamFusion \cite{poole2022dreamfusion} employs Imagen \cite{saharia2022photorealistic} for Score Distillation Sampling (SDS) to supervise the generation of NeRF \cite{mildenhall2021nerf} models. Although diffusion-based methods generate more impressive content, they often fail to converge when generating human bodies and cannot control the generated results even when they do. Therefore, using diffusion models to supervise the generation of 3D avatars remains a challenge.

\noindent\textbf{Clothing-decoupled models.}
Clothing-body separation was initially developed for simulating clothing in physics simulations \cite{santesteban2019learning, bertiche2020cloth3d, patel2020tailornet, vidaurre2020fully}. Recently, clothing-body separation models \cite{jiang2020bcnet, zhu2020deep, feng2022capturing, corona2021smplicit} have emerged in avatar generation and reconstruction tasks. These tasks are more challenging because separating 3D clothing and body purely from 2D data is difficult and requires extensive pixel-level annotation of clothing and body data.
SMPLicit \cite{corona2021smplicit} is a generative clothing-decoupled model that uses a neural implicit network to represent clothes outside the SMPL mesh. Neural implicit models can represent clothing with any topology, thus significantly enhancing the expressive power compared with previous complete mesh-based models. 
The task AvatarFusion faces is even more challenging as there is no ground truth.
\begin{figure*}[h]
\centering
    \includegraphics[scale=0.3]{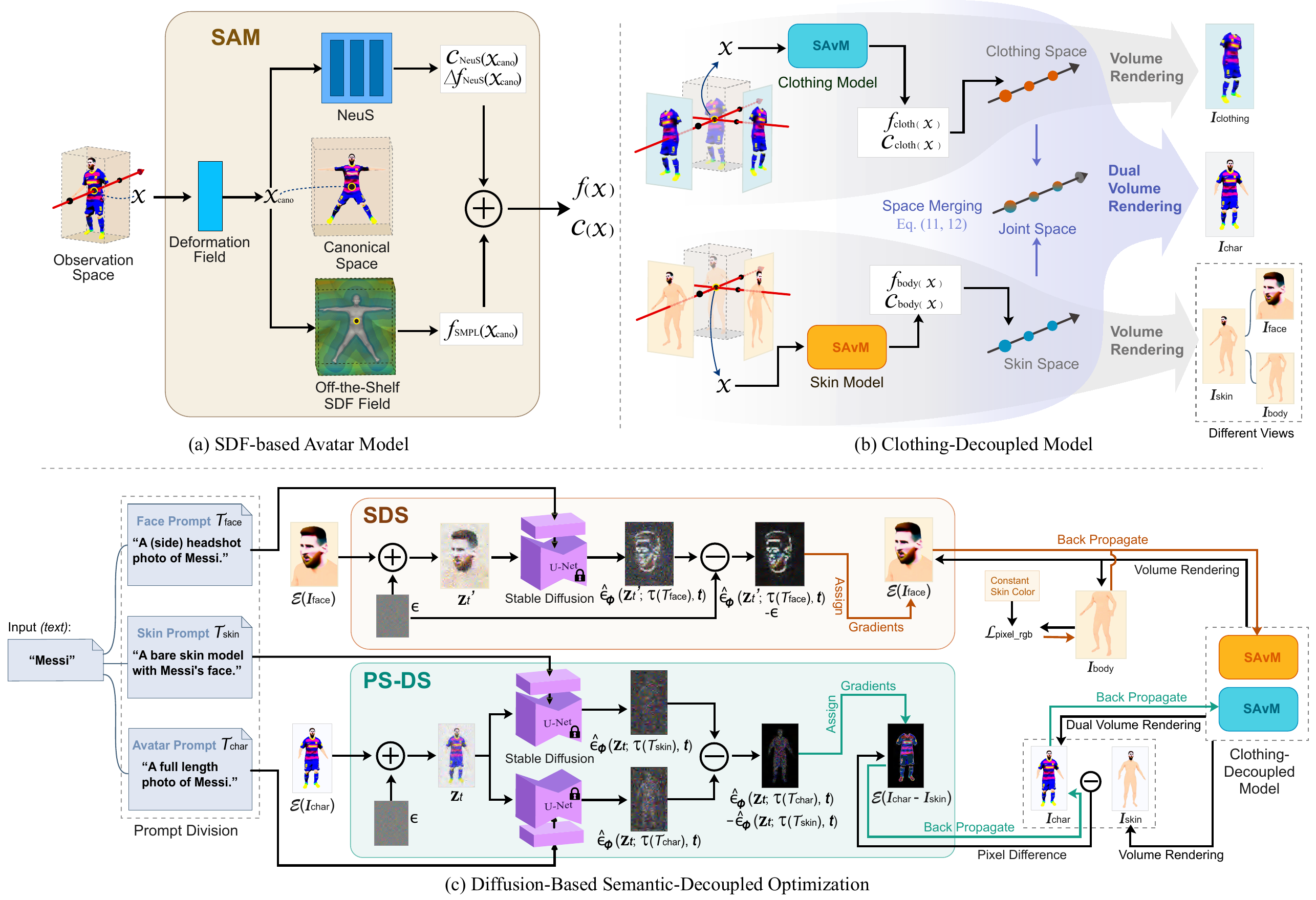}
    \caption{Overview of AvatarFusion. The upper left part shows (a) the SDF-Based Avatar Model (SAvM) which takes a point $\mathbf{x}$ as input, and output its SDF value and color value. The upper right part shows (b) the clothing-decoupled model, which takes two SAvMs representing skin and clothing, and merge the space to render avatars with clothes. The lower part shows (c) our diffusion-based optimization methods with PS-DS separating the clothing from skin semantically. For clarity, we omit the image encoder $\mathcal{E}$ in the figure.}
    \label{fig:pipeline}
\end{figure*}

\section{Preliminaries}
\textbf{Human body priors.}
SMPL (Skinned Multi-Person Linear) \cite{loper2015smpl} is a statistical body model that represents the surface of the human body as a triangulated mesh $M(\beta, \theta)$ with vertices $\mathbf{v}_i$. The model uses shape parameters $\beta$ and pose parameters $\theta$ to control the body's shape and pose, respectively. To enable deformation of the mesh based on pose, SMPL follows a linear blending skinning function for vertex transformations as:

\begin{equation} \label{equ:smplrotation}
    \mathbf{v}_i^{\prime} = \sum_{k=1}^{K}{\omega_{k,i}}G^{\prime}_{k}(\theta, \mathbf{J})\mathbf{v}_i,
\end{equation}
where $\omega_{k,i}$ are the blend weights for the $k$-th joint, and $\mathbf{G}^{\prime}_k(\theta, \mathbf{J})$ is the relative transformation matrix of the $k$-th joint dependent on the pose parameter $\theta$ and joint locations $\mathbf{J}$. 

\noindent\textbf{Neural implicit surfaces.}
Neural Implicit Surfaces (NeuS) \cite{wang2021neus} is a recently developed method for learning implicit representations of 3D surfaces from 2D images. The model uses the volume rendering \cite{drebin1988volume, mildenhall2021nerf}, a technique to project a 3D volume onto a 2D image plane, to train a neural network to reconstruct the 3D surface as Signed Distance Function (SDF). The volume rendering strategy estimates the pixel color $\hat{C}$ by shooting a ray $\mathbf{p}(t)$ from the camera and calculating properties of sampled points $\mathbf{p}(t_i)$ on the ray, using three primary equations:
\begin{equation} \label{equ:neusalpha}
\alpha_{i}=1-\mathrm{exp}(-\int^{t_{i+1}}_{t_{i}} {\rho(t)\mathrm{d}t}),
\end{equation}
\begin{equation} \label{equ:neust}
T_{i}=\prod^{i-1}_{j=1} {(1 - \alpha_{j})},
\end{equation}
\begin{equation} \label{equ:volumerendering}
\hat{C}=\sum^{n}_{i=1} {T_{i} \alpha_{i} \mathbf{c}_{i}}.
\end{equation}
Here $\alpha_i$ represents the discrete transparency of each point, which is calculated as the integral of the density function $\rho(t)$ between neighbour points. $T_i$ represents the the accumulated transparency of all points behind it along the viewing ray. The pixel color $\hat{C}$ is then calculated based on $\alpha_i$, $T_i$, and point color $\mathbf{c}_i$.

\noindent\textbf{Diffusion-supervised training.}
DreamFusion \cite{poole2022dreamfusion} proposes a method for generating 3D models from text using a pre-trained diffusion model called Imagen \cite{saharia2022photorealistic} with a denoising function, denoted as $\hat{\epsilon}_{\phi}(\mathbf{z}_t; y, t)$, where $\mathbf{z}_t$ represents a noisy image at noise level $t$, and $y$ is the text embedding. The optimization of the neural implicit model $g$ begins by adding noise $\epsilon$ to the rendered image $\mathbf{x}=g(\theta)$ at a specified noise level $t$, obtaining $\mathbf{z}_t$. The Score Distillation Sampling (SDS) technique is then employed to optimize the model towards the intended text meaning by calculating the difference between the predicted text-conditioned noise, $\hat{\epsilon}_{\phi}(\mathbf{z}_t; y, t)$, and the added noise $\epsilon$, as follows:
\begin{equation}
\label{equ:SDS}
    \nabla_{\theta}{\mathcal{L}_{\mathrm{SDS}}}{(\phi, \mathbf{x}=g(\theta))} \triangleq \mathbb{E}_{t,\epsilon}[\omega(t)(\hat{\epsilon}_{\phi}(\mathbf{z}_{t};y,t)-\epsilon) \frac{\partial\mathbf{x}}{\partial\theta}],
\end{equation}
where $\omega(t)$ is a weighting function.

\section{Methodology}
\subsection{Overview}

In this study, we present AvatarFusion, a novel framework designed to generate photo-realistic avatars with separate clothing for the zero-shot text-to-avatar task. Figure \ref{fig:pipeline} illustrates the pipeline of AvatarFusion. Our framework incorporates a Clothing-Decoupled neural implicit Model (CDM) which consists of two SDF-based Avatar Models (SAvM), one for skin and the other for clothing. To optimize the models, we utilize Diffusion-Based Supervisions.

We begin by introducing the SAvM in Section \ref{subsection:BaseModel}. The SAvM extends NeuS \cite{wang2021neus} by incorporating a deformation field and an off-the-shelf SDF field generated by an offline SDF generator. The deformation field transforms points in an observation space to points in the canonical space. Meanwhile, the SDF field provides a detailed human body prior, SMPL \cite{loper2015smpl}, enabling our model to render well-formed facial structures and detailed SMPL models before any optimization, which effectively mitigates facial distortion problems.

In Section \ref{subsection:Clothing-DecoupledModel}, we provide detailed explanations of CDM. CDM is our full model that represents the skin and clothing of an avatar using two SAvMs. These SAvMs can produce skin and clothing images using SDF-based volume rendering \cite{wang2021neus}. To render avatars with clothing, we merge the space points from two neural implicit representations using Dual Volume Rendering.

In Section \ref{subsection:Optimization}, we describe the process of optimizing the models using Stable Diffusion, a latent diffusion model \cite{rombach2022high}.
Firstly, we divide the text description into three prompts: face, skin, and avatar prompt. The face prompt is used as the text condition of the SDS method \cite{poole2022dreamfusion} to optimize the facial part of the skin model. For the body part of the skin model, we use a selected skin color. The skin and avatar prompts are regarded as two text conditions for Pixel-Semantic Difference-Sampling (PS-DS) to generate clothes, as the overall avatar minus the bare skin leaves only the clothes.

\subsection{SDF-Based Avatar-Specific Model} \label{subsection:BaseModel}
NeuS has the ability to produce high-precision surface reconstructions for static scenes. However, in order to better represent an avatar, we have extended NeuS to the SAvM by introducing a deformation field and an off-the-shelf SDF field.

\noindent\textbf{Deformation field.}
To create a dynamic avatar, we require a deformation field that maps the observation space, where the avatar takes on arbitrary poses $\theta$, to a canonical space, where the avatar is in a standard pose $\theta_{\mathrm{cano}}$. Following the approach of \cite{peng2021animatable, park2021nerfies, pumarola2021d, pan2023transhuman}, we align any given point $\mathbf{x}$ in the observation space to the nearest vertex $\mathbf{v}$ of SMPL mesh $M(\beta, \theta)$ based on Euclidean distance, and assign the blend weights of $\mathbf{v}$, denoted as $\omega_{i}$, to $\mathbf{x}$. This allows $\mathbf{x}$ to rotate with SMPL joints and correspond to a point $\mathbf{x}_{\mathrm{cano}}$ in the canonical space, following Equation \ref{equ:smplrotation}. The deformation field equation is then defined as follows:

\begin{equation}
    \mathbf{x}_{\mathrm{cano}}=\sum_{k=1}^{K}\omega_{k}\mathbf{G}^{\prime}_{k}(\theta, \mathbf{J})\mathbf{x}.
\end{equation}

\noindent\textbf{Off-the-shelf SDF field.}
The optimization process begins with the models representing SMPL prior, as a starting point \cite{michel2022text2mesh, hong2022avatarclip}. To encode the SMPL prior directly into the model, we use an offline SDF generator to create a $256\times256\times256$ grid of SDF values that correspond to the SMPL mesh space at its canonical pose $M(\beta, \theta_{\mathrm{cano}})$. Any point $\mathbf{x}_{\mathrm{cano}}$ in the mesh space can be assigned an SDF value denoted as $f_{SMPL}(\mathbf{x}_{\mathrm{cano}})$ using bilinear interpolation.
The Multi-Layer Perceptrons (MLPs) of NeuS generate only a residual SDF value $\Delta f_{NeuS}(\mathbf{x}_{\mathrm{cano}})$. This ensures that the model can render highly detailed SMPL even before any training has been conducted. 

\noindent\textbf{Summary} Therefore, the SDF value $f(\mathbf{x})$ and color $\mathbf{c}(\mathbf{x})$ at point $\mathbf{x}$ for volume rendering Equation \ref{equ:volumerendering} is
\begin{equation}
    \Delta f_\mathrm{NeuS}(\mathbf{x}_{\mathrm{cano}}), \mathbf{feat}=MLPs(\mathbf{x}_\mathrm{cano}),
\end{equation}
\begin{equation}
    \mathbf{c}(\mathbf{x})=\mathbf{c}_\mathrm{NeuS}(\mathbf{x}_\mathrm{cano})=MLPs(\mathbf{x}_{\mathrm{cano}}, \mathbf{feat}),
\end{equation}
\begin{equation}
    f(\mathbf{x})=f_\mathrm{SMPL}(\mathbf{x}_{\mathrm{cano}})+\Delta f_\mathrm{NeuS}(\mathbf{x}_{\mathrm{cano}}).
\end{equation}

\begin{figure*}[h]
\centering
    \includegraphics[scale=0.3]{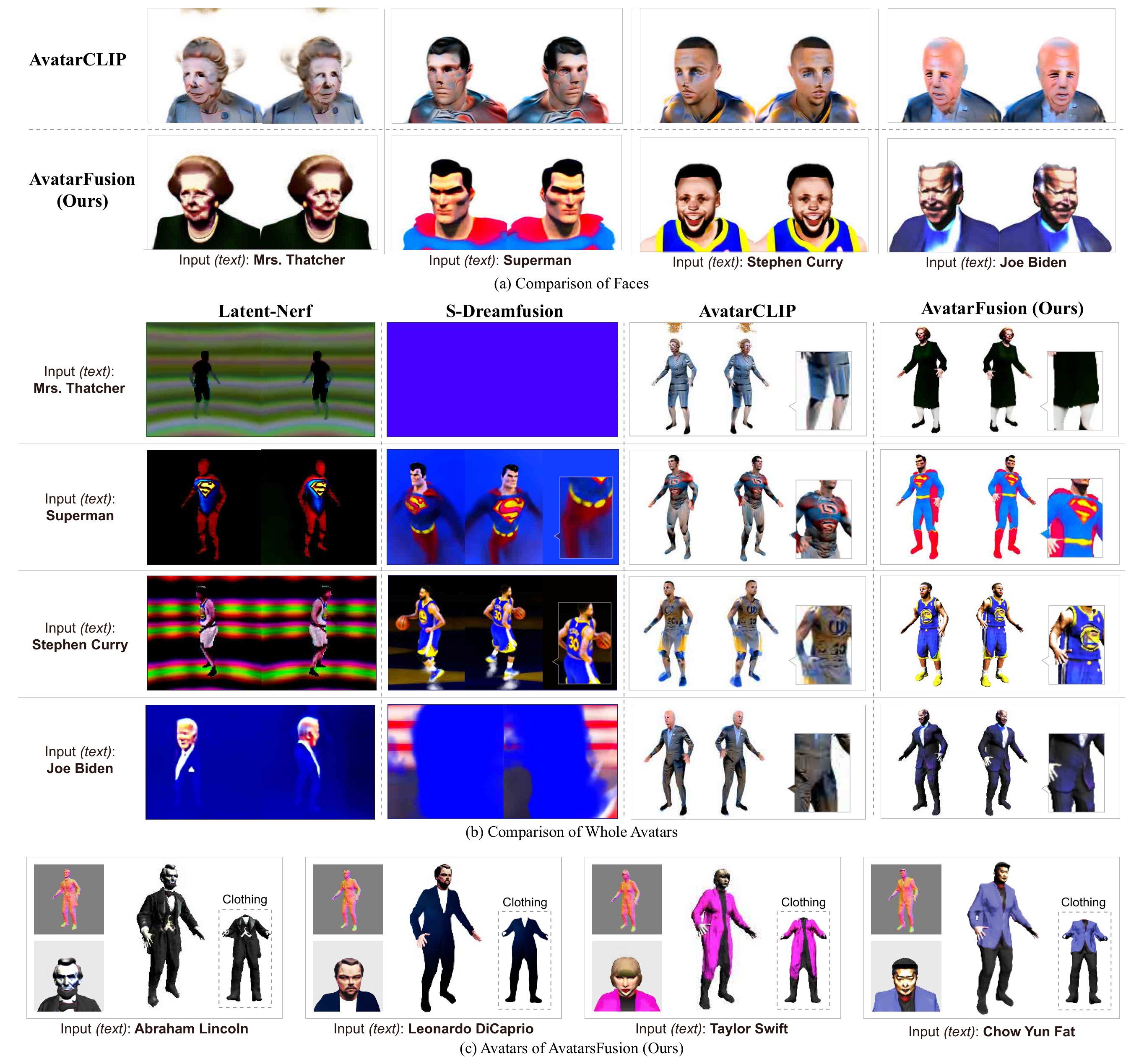}
    \caption{Qulitative Comparison with baselines for face and body generation and more results of AvatarFusion. Latent-NeRF and Stable-DreamFusion can sometimes fail to generate content or produce distorted bodies. On the other hand, AvatarCLIP suffers from poor details. In comparison, our method can robustly generate avatars with vivid faces and realistic clothing. More examples are provided in the supplementary material.}
    \label{fig:baseline}
\end{figure*}

\subsection{Clothing-Decoupled Model} \label{subsection:Clothing-DecoupledModel}
\noindent\textbf{Skin, clothing and joint spaces.}
Our goal is to create a more realistic clothing simulation by developing a clothing-decoupled neural implicit model that represents skin and clothes separately. To achieve this, we utilize two SAvMs, one for rendering the images of skin space $I_\mathrm{skin}$ and the other for rendering the clothing space $I_\mathrm{clothing}$. However, both SAvMs must be integrated into a joint space to render the complete image $I_\mathrm{char}$. One direct approach involves dividing the joint space into an inner space, where only the skin space points exists, and an outer space, where only the clothing space points exists, as proposed in \cite{feng2022capturing}.
\begin{equation}
\begin{split}
    \mathbf{x} = \left\{
    \begin{array}{lr}
    \mathbf{x}_{\mathrm{body}},   &  f_\mathrm{body}(\mathbf{x}) <= \delta \\
    \mathbf{x}_{\mathrm{cloth}},   &  f_\mathrm{body}(\mathbf{x}) > \delta 
    \end{array}
    \right.,
\end{split}
\end{equation}
where $\delta$ is a small positive number.

However, connecting the skin and clothing spaces directly to form the joint space may result in a biased skin color that differs from the skin color rendered in the skin space. This is because skin color is determined by integrating every point's color along the ray during volume rendering. When skin space points are neglected in the outer joint space, their skin color contribution may also be disregarded, leading to a biased skin color. Therefore, the skin space points must also exist in the outer joint space to maintain their distribution.

\noindent\textbf{Dual Volume Rendering.}
To integrate the two neural implicit models in the outer joint space, an extended volume rendering technique called Dual Volume Rendering is proposed. Specifically, we update the properties of joint space points, $\alpha_{i}$ and $\mathbf{c}_{i}$ in Equation \ref{equ:volumerendering}, with combined values from both models. To be concise, we provide the summarized space merging equations below and leave the detailed derivation process in the supplementary material. Please note that the derivation of the merged $\alpha_i$ is in line with the definition of volume rendering \cite{drebin1988volume}, while the merged $\mathbf{c}_i$ is a linear interpolation approximation as there is no physical meaning of adding the RGB values.

\noindent \textbf{Space merging equations.} Given the sampled points on a ray, denoted as $\mathbf{X}=\{\mathbf{x}_1, \mathbf{x}_2, \dots, \mathbf{x}_i, \dots\}$, we derive $f_\mathrm{body}(\mathbf{x}_i)$, $f_\mathrm{cloth}(\mathbf{x}_i)$, $\mathbf{c}_\mathrm{body}(\mathbf{x}_i)$, and $\mathbf{c}_\mathrm{cloth}(\mathbf{x}_i)$ from the SAvMs. Then, we calculate $\alpha_\mathrm{body}(\mathbf{x}_i)$ and $\alpha_\mathrm{cloth}(\mathbf{x}_i)$ based on the discretization of Equation \ref{equ:neusalpha} (refer to \cite{wang2021neus} for more details). Finally, Dual Volume Rendering procedure can be summarised as:
\begin{equation}
\begin{split}
    \alpha_i = \left\{
    \begin{array}{ll}
    \alpha_\mathrm{body}(\mathbf{x}_i),   &  f_\mathrm{body}(\mathbf{x}_i) <= \delta \\
    \alpha_\mathrm{body}(\mathbf{x}_i) + \alpha_\mathrm{cloth}(\mathbf{x}_i) \\ - \alpha_\mathrm{body}(\mathbf{x}_i) \cdot \alpha_\mathrm{cloth}(\mathbf{x}_i),   &  f_\mathrm{body}(\mathbf{x}_i) > \delta 
    \end{array},
    \right.
\end{split}
\end{equation}

\begin{equation}
\begin{split}
    \mathbf{c}_i = \left\{
    \begin{array}{ll}
    \mathbf{c}_\mathrm{body}(\mathbf{x}_i),   &  f_\mathrm{body}(\mathbf{x}_i) <= \delta \\
    \frac{\alpha_\mathrm{body}(\mathbf{x}_i)}{\alpha_\mathrm{body}(\mathbf{x}_i) + \alpha_\mathrm{cloth}(\mathbf{x}_i)} \mathbf{c}_\mathrm{body}(\mathbf{x}_i) \\ + \frac{\alpha_\mathrm{cloth}(\mathbf{x}_i)}{\alpha_\mathrm{body}(\mathbf{x}_i) + \alpha_\mathrm{cloth}(\mathbf{x}_i)}\mathbf{c}_\mathrm{cloth}(\mathbf{x}_i) ,   &  f_\mathrm{body}(\mathbf{x}_i) > \delta 
    \end{array}.
    \right.
\end{split}
\end{equation}
Hereafter, we can proceed with the subsequent volume rendering calculations using Equation \ref{equ:neust} and so forth.

\subsection{Diffusion-Based Semantic-Decoupled Optimization}
\label{subsection:Optimization}
\textbf{Prompt division.}
In this sub-section, we present our optimization methods using Stable Diffusion for generating a complete avatar with decoupled skin and clothing components. To achieve this, we generate three prompts: the face prompt $\mathcal{T}_\mathrm{face}$, the skin prompt $\mathcal{T}_\mathrm{skin}$, and the character prompt $\mathcal{T}_\mathrm{char}$, which correspond to the avatar's facial features, skin tone, and overall appearance, respectively.

We use $\mathcal{T}_\mathrm{face}$ as the text condition for the SDS method \cite{poole2022dreamfusion} to optimize the facial part of skin model. We use a pre-selected color to optimize $I_\mathrm{body}$, the body image rendered in the skin space, with pixel-wise loss $\mathcal{L}_\mathrm{pixel\_rgb}$. $\mathcal{T}_\mathrm{skin}$ and $\mathcal{T}_\mathrm{char}$ are used to generate the clothing of avatar indirectly. This is because the diffusion model can only generate clothing along with skin, and directly applying the SDS method with a clothing prompt as a condition would result in a new layer of skin outside the original skin. To address this issue, we introduce the Pixel-Semantic Difference-Sampling (PS-DS) method.

\noindent\textbf{PS-DS method.}
This method takes two text prompts as input and generates semantic differences between them. By using $\mathcal{T}_\mathrm{skin}$ and $\mathcal{T}_\mathrm{char}$ as input, we obtain clothing as their semantic difference.

The PS-DS method is implemented as shown in Figure \ref{fig:pipeline}.
First, we render two images $I_\mathrm{skin}$ and $I_\mathrm{char}$ and encode $I_\mathrm{char}$ to its latent image $\mathcal{E}(I_\mathrm{char})$ using Stable Diffusion's encoder. Then, noise $\epsilon$ at level $t$ is added to $\mathcal{E}(I_\mathrm{char})$ to obtain $\mathbf{z}_t$, which serves as a noisy sample for Stable Diffusion.
Next, we predict noise based on $\mathcal{T}_\mathrm{char}$ and $\mathcal{T}_\mathrm{skin}$, respectively. The difference between the predicted noises provides optimizing directions for the clothing model. We compute these directions using the equation below:
\begin{equation}
    \begin{split}
        \nabla_{\theta}{\mathcal{L}_{\mathrm{PS\mbox{-} DS}}}{(\phi, \mathbf{x}=g(\theta))} \triangleq \mathbb{E}_{t,\epsilon}[\omega(t)(\hat{\epsilon}_{\phi}(\mathbf{z}_t;\tau(\mathcal{T}_\mathrm{char}),t) \\ -\hat{\epsilon}_{\phi}(\mathbf{z}_t;\tau(\mathcal{T}_\mathrm{skin}), t)) \frac{\partial\mathrm{x}}{\partial\theta}],
    \end{split}
\end{equation}
where $\tau$ represents the text encoder of Stable Diffusion. For definitions of other variables, please refer to Equation \ref{equ:SDS}.
We assign this result as the gradients to the latent image $\mathcal{E}(I_\mathrm{char}-I_\mathrm{skin})$ and backpropagate it.

In addition to SDS and PS-DS method, we also use SDF loss to control the overall shape of the model and the pixel entropy loss to assist PS-DS method in segmenting the clothing. By calculating the entropy of the proportion of skin and clothing model in the color of each ray as a loss function, one side can dominate the overall color, which can better distinguish whether there is clothing coverage.

\section{Experiments}
\subsection{Implementation Details}
We represented the SAvMs of the body and clothing using two-layer MLPs with a hidden size of 128. Our latent diffusion model is the open-source Stable Diffusion model version 1.5. We trained the whole avatar in a standing pose as in AvatarCLIP \cite{hong2022avatarclip}. We trained the skin model for 25000 iterations and then used PS-DS to train the clothing model for 25000 iterations with $\delta=1e-5$, followed by an additional 25000 iterations with $ \delta=1e-3$. More details are provided in the supplementary material.

\subsection{Benchmark}
\begin{figure}[t]
\centering
    \includegraphics[scale=0.18]{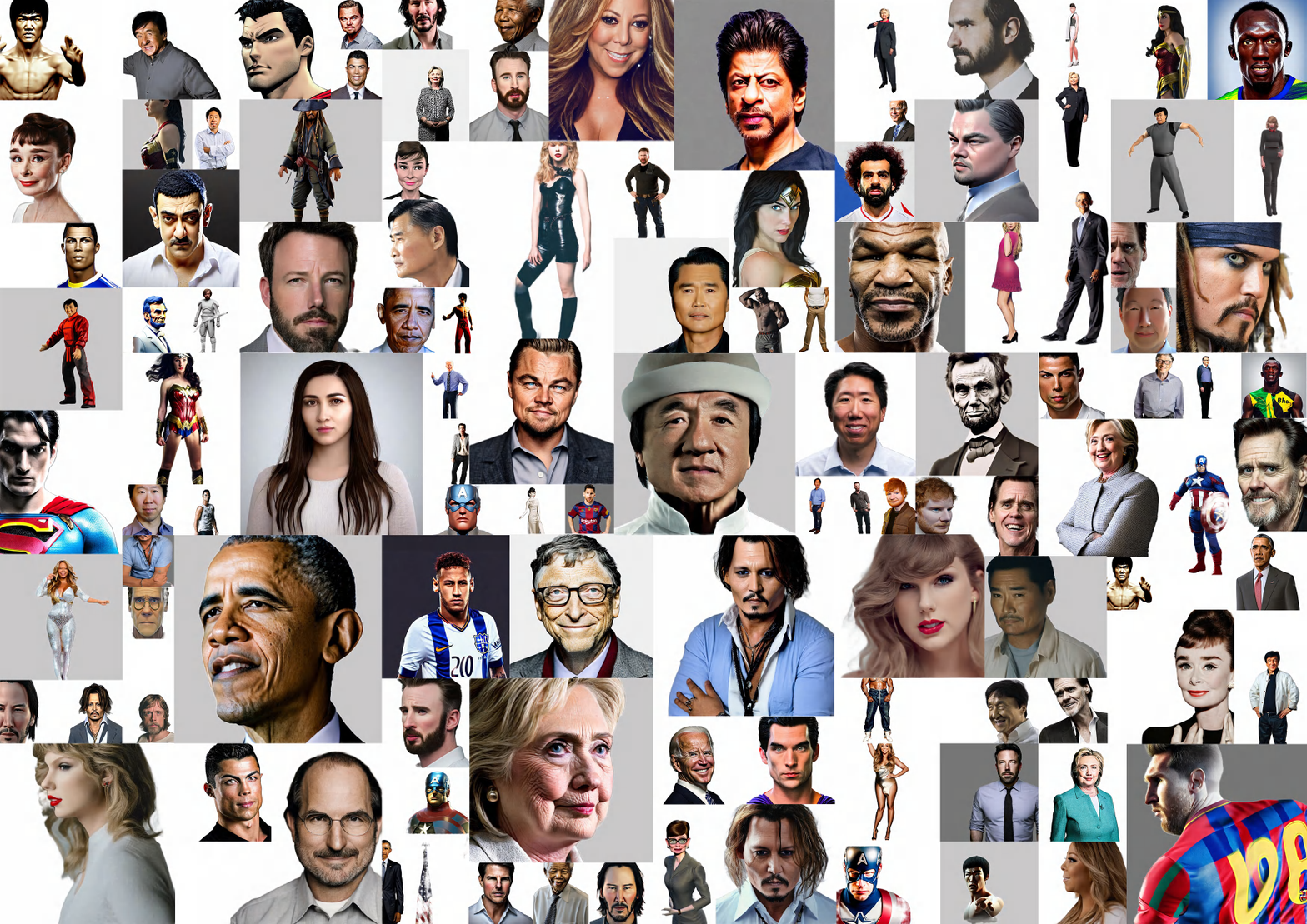}
    \caption{The constructed dataset of the benchmark.}
    \label{fig:dataset}
\end{figure}

The evaluation of zero-shot text-to-avatar generation task is challenging due to the absence of a unified dataset of selected characters and evaluation metrics beyond user studies. To address this, we developed a benchmark called Famous-Character-50 (FC50) that includes 50 descriptions of famous real and fictional characters, and the corresponding images generated using the Stable Diffusion \cite{rombach2022high} checkpoints of version 2.1. FC50 provides a diverse representation of races and genders while covering a broad range of cultures, as shown in Figure \ref{fig:dataset}. This diversity guarantees a comprehensive evaluation that accurately reflects the diversity of human appearance.

\noindent\textbf{Evaluation metrics.}
Since there is no ground truth available for the zero-shot text-to-avatar task, we propose comparing the similarity between 3D model rendering images and mature 2D image generation results to evaluate the models. To measure this similarity, we employ a face recognition model \cite{geitgey2017face} to capture and measure the distance between the generated face and the corresponding face in the image dataset, which we refer to as the Face Recognition Distance (FRD). A smaller distance indicates that the generated avatar is closer to the corresponding face in terms of identity recognition, which suggests that the avatar better captures the character features described in the text prompt. Additionally, we compare the FID score between the generated avatars and image dataset, evaluating the quality of the generated avatars in terms of fidelity to the ground truth images. We compare the faces (Face-FID) and the whole body (Body-FID) separately.

\begin{table}[t]
\centering
\caption{Comparison with baselines and ablation studies. For all the metrics, the lower (↓), the better.}
\begin{tabular}{l c c c c c}\hline
Model       &FRD  &Face-FID   &Body-FID\\\hline
Latent-NeRF \cite{metzer2022latent} & 0.8956       &  105.87      &114.73\\
S-DreamFusion \cite{stable-dreamfusion} &  (-)      &  (-)      &119.59\\
AvatarCLIP \cite{hong2022avatarclip}  &0.6907     &79.08      &96.65\\
Ours        &\textbf{0.5937}     &\textbf{71.73}      &\textbf{83.21}\\
\hline
Full model      & 0.6007 & 82.13  & 92.16\\
32 * 32 SDF Reso& 0.7581 & 104.69 & 95.42\\
64 * 64 SDF Reso& 0.6754 & 87.94  & 96.85\\
w/o diffusion   & 0.7023 & 96.08  & 104.32\\
w/o speartaion  & 0.6018 & 84.72  & 95.98\\

    \hline
\end{tabular}
\label{tab:baseline}
\end{table}

\subsection{Comparing with Baselines}

We compared AvatarFusion with three baseline methods: DreamFusion \cite{poole2022dreamfusion}, Latent-NeRF \cite{metzer2022latent}, and AvatarCLIP \cite{hong2022avatarclip}. 
DreamFusion is a generative model that converts text to 3D objects using Imagen, a diffusion model \cite{saharia2022photorealistic}. As we were unable to obtain Imagen, we used an unofficial implementation called Stable-DreamFusion (S-DreamFusion) \cite{stable-dreamfusion}, which is based on Stable Diffusion \cite{rombach2022high}. Latent-NeRF modifies the generation space of S-DreamFusion from RGB space to the latent space of Stable Diffusion and uses a loss function to align the 3D implicit representation to a mesh prior representing basic shapes. We used SMPL \cite{loper2015smpl} prior for avatar generation in this method. AvatarCLIP is specifically designed for avatar generation, and its robustness and better combination with the SMPL prior make it a strong competitor. For AvatarCLIP, we used the official text prompt "A 3D rendering of {character's name} in Unreal Engine," while for methods utilizing diffusion models, we used the text prompt "A full-length photograph of {character's name}."


\noindent \textbf{Quantitative comparison.}
Table \ref{tab:baseline} shows the evaluation results of our model against baselines.
Our superior performance on all three evaluation metrics indicates that we are capable of generating more accurate and visually appealing avatars from textual descriptions. Specifically, our models achieve a face distance of 0.5937, which is 14.04\% higher than the baseline, indicating our generated faces are more realistic and faithful to the text.

\noindent\textbf{Qualitative comparison.}
The generation of complex human bodies using diffusion-based methods, such as Latent-NeRF and S-DreamFusion, is not always successful. S-DreamFusion failed to generate meaningful content for Mrs. Thatcher and Joe Biden, while Latent-NeRF initially converged but later lost all content. We present the best results before this issue occurred. While S-DreamFusion can generate avatars in some cases, it falls short in terms of facial and clothing details compared with AvatarFusion. Additionally, it can cause body distortions, such as generating three arms for Curry or not producing hands for Superman.
These problems of diffusion-based method arise because of the models' ineffective combination with SMPL prior, which results in rendered images with weak correlations across different perspectives. Applying diffusion-based pixel-level supervision to different perspectives at this stage may lead to convergence failure or distortions.

Both AvatarCLIP and AvatarFusion can robustly generate complete avatars. However, the generic model of AvatarCLIP and the embedding-level supervision of CLIP cannot distinguish between clothing and skin, which results in characters with distorted faces, mixed clothing and skin textures, and clothing lacking thickness.

\begin{table}[t]
\centering
\caption{Results of user studies from 19 users. All metrics are measured on a scale of 1 (worst) to 5 (best). AvatarFusion achieves highest rates on all aspects.}
\begin{tabular}{l c c c c c c}\hline
Model&Overall&Face&Texture&Text Cons\\\hline
Latent-NeRF \cite{metzer2022latent} & 2.21 & 1.89 & 2.47 & 2.73\\
S-DreamFusion \cite{stable-dreamfusion} & 3.74 & (-) & 3.95 & 2.84\\
AvatarCLIP \cite{hong2022avatarclip}  & 3.89 & 3.00 & 4.16 & 4.58\\
Ours & \textbf{4.32} & \textbf{4.68} & \textbf{4.42} & \textbf{4.79}\\
    \hline
\end{tabular}
\label{tab:userstudy}
\end{table}

\noindent\textbf{User studies.}
To further evaluate the quality of our generated avatars, we conducted a user study comparing them with baseline methods. We recruited 19 volunteers and asked them to rate the methods based on (1) overall quality, (2) facial quality, (3) texture quality, and (4) consistency with the given text prompt. For each aspect, we randomly selected 12 generated results and asked the volunteers to score each example on a scale of 1 (worst) to 5 (best). The final results are presented in Table \ref{tab:userstudy}. Our AvatarFusion approach achieved the highest rank in all aspects, demonstrating its effectiveness in mitigating the defects of previous methods.

\begin{figure}[t]
\centering
    \includegraphics[scale=0.32]{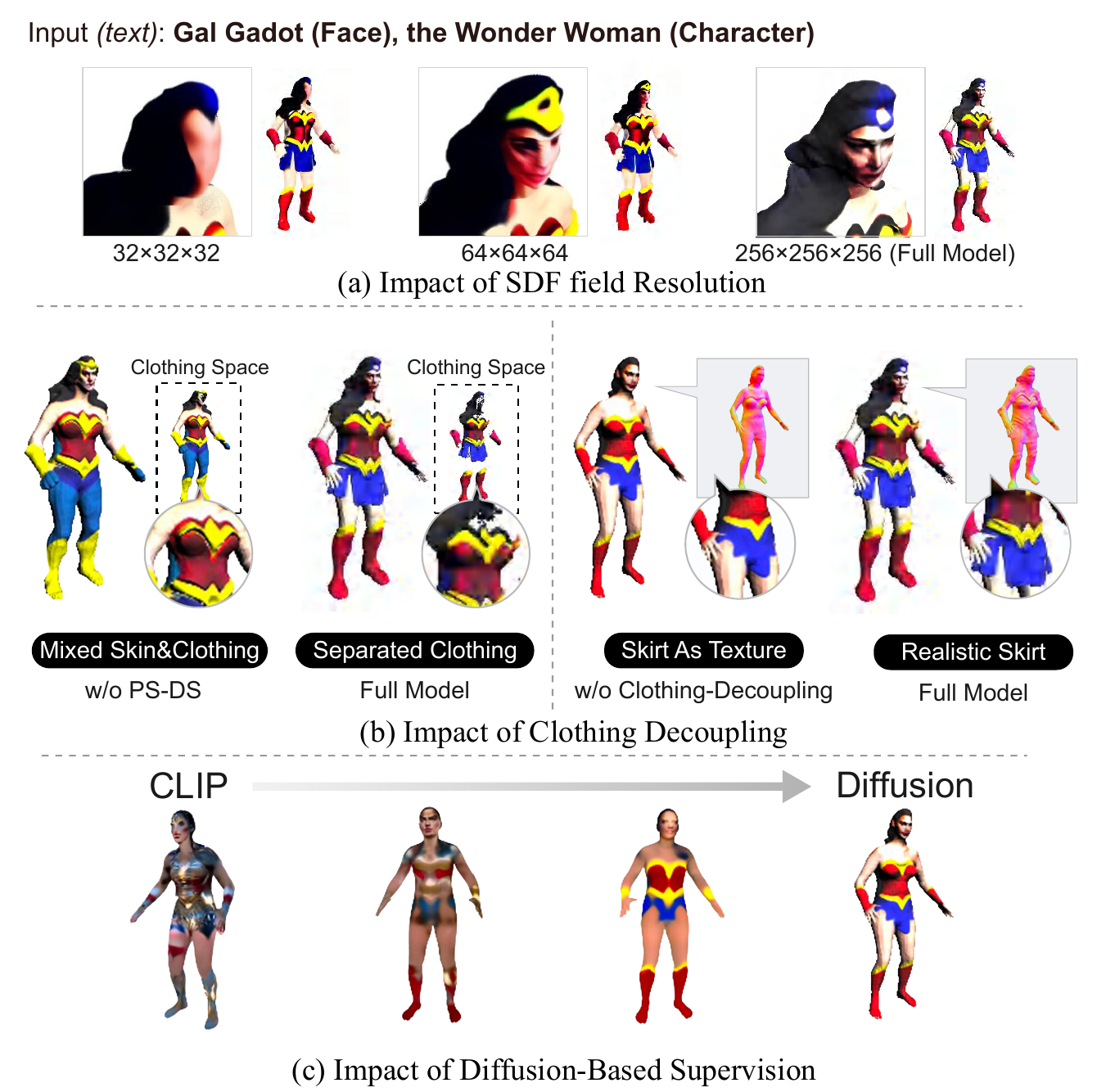}
    \caption{Ablation studies which demonstrate the following key findings: (a) High-resolution SDF fields contribute to well-formed faces. (b) The semantic separation of clothing from skin by PS-DS method allows for more realistic skirts than just texture alone. (c) The inclusion of Diffusion Supervision contributes to a more complete and detailed clothing.}
    \label{fig:ablation}
\end{figure}

\subsection{Ablation Studies}

We conducted ablation studies on a random sample of fifteen characters from FC50 to gain insight into the contributions of each component to the overall performance. The results of these studies are presented in Figure \ref{fig:ablation} and Table \ref{tab:baseline}.

\noindent\textbf{Impact of SDF field resolution.}
We examined the impact of using off-the-shelf SDF fields of different resolutions on our framework's performance. Our full model used a $256^3$ SDF grid. Since there was no noticeable difference from the human perspective when using a $128^3$ grid, we evaluated the results for SDF grids with resolutions of $32^3$ and $64^3$. The results showed that when we reduced the resolution, the avatar model at initial point lacked the necessary details, which ultimately led to facial distortion in the final output.

\noindent\textbf{Impact of clothing decoupling.}
We investigated the impact of clothing decoupling on the results. First, we replaced PS-DS method with SDS method \cite{poole2022dreamfusion}, which resulted in a loss of semantic decoupling between the Wonder Woman's skin and clothing in Figure \ref{fig:ablation}. Then, we compared the results obtained with and without clothing decoupling and found that the decoupled model produced more expressive skirts than just a texture on the legs. Additionally, it achieved better performance in evaluation metrics. Decoupling clothing from skin enabled independent optimizing of clothing, loosening the constraints of the SMPL model, a body prior rather than a clothing prior.

\noindent\textbf{Impact of diffusion-based supervision.}
We conducted an experiment to evaluate the impact of diffusion-based supervision by replacing it with CLIP \cite{radford2021learning}. In this step, we also removed clothing decoupling as it relies on diffusion supervision. The results demonstrated that CLIP-based optimization led to a blending of skin and clothing textures, whereas diffusion supervision resulted in complete clothing textures.

\begin{figure}[t]
\centering
    \includegraphics[scale=0.29]{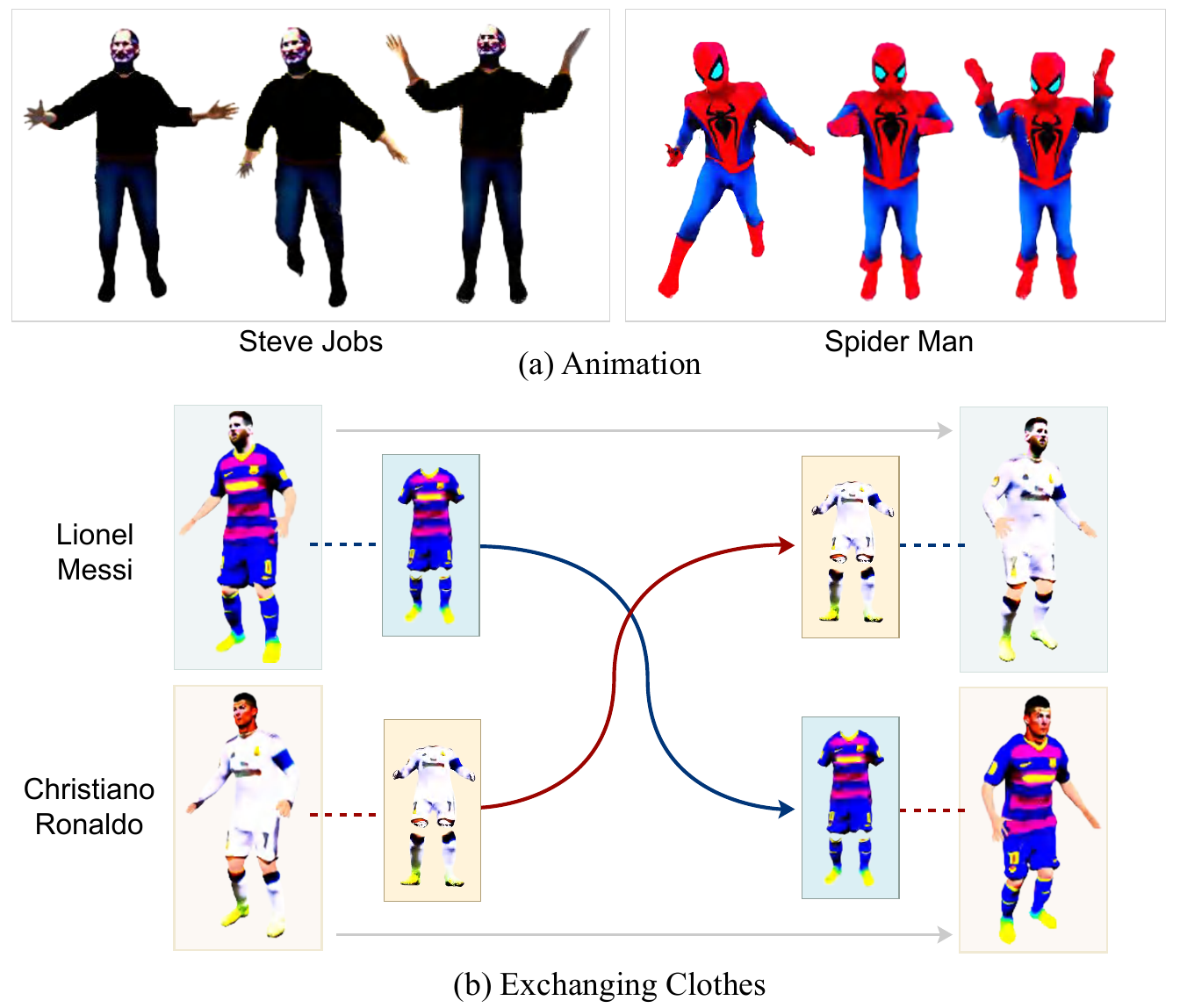}    \caption{Results of (a) animation and (b) exchanging clothes.}
    \label{fig:exchange}
\end{figure}
\subsection{Extra Abilities}
As we align the generated avatars with the SMPL skeleton, they can be animated with SMPL pose sequences. And as our avatar is clothing-decoupled, we can exchange the clothes of avatars as shown in Figure \ref{fig:exchange}.
\section{Discussion}
\textbf{Limitations.}
Similar to AvatarCLIP \cite{hong2022avatarclip}, our method currently cannot generate a reasonable backside because the vision-language models are not responsive to the prompt "the back of ...". Consequently, the backside of the avatar may resemble the front.

\noindent\textbf{Ethical issues.}
The technology of generating realistic and clothing-decoupled avatars from text may raise concerns regarding ethics, privacy, and security. It is critical to continue conducting research and development in an ethical and responsible manner. Please note that in our avatar generation process, we only generate the inner model representing skin color and not the specific physique of an individual. We firmly oppose generating unethical outcomes.

\section{Conclusions}
In this study, we introduce AvatarFusion, a novel framework for zero-shot text-to-avatars generation. Our primary contribution is a clothing-decoupled neural implicit avatar model that employs a dual volume rendering strategy and a Pixel-Semantics Difference-Sampling process to separate the generation of body and clothing. Our experiments on the first benchmark in this field demonstrate that our approach surpasses state-of-the-art methods by a significant margin. We are the first to generate and segment 3D avatar models using a pre-trained vision-language model. Future research may focus on enhancing the backside of avatars or fitting loose clothing.

\section*{Acknowledgments}
This work is supported by the National Key R\&D Program of China under Grant No. 2021QY1500, the State Key Program of the National Natural Science Foundation of China (NSFC) (No. 61831022), the Fundamental Research Funds for the Central Universities (No. 226-2022-00051).

\clearpage
\bibliographystyle{ACM-Reference-Format}
\balance
\bibliography{acmart}


\begin{thebibliography}{56}


\ifx \showCODEN    \undefined \def \showCODEN     #1{\unskip}     \fi
\ifx \showDOI      \undefined \def \showDOI       #1{#1}\fi
\ifx \showISBNx    \undefined \def \showISBNx     #1{\unskip}     \fi
\ifx \showISBNxiii \undefined \def \showISBNxiii  #1{\unskip}     \fi
\ifx \showISSN     \undefined \def \showISSN      #1{\unskip}     \fi
\ifx \showLCCN     \undefined \def \showLCCN      #1{\unskip}     \fi
\ifx \shownote     \undefined \def \shownote      #1{#1}          \fi
\ifx \showarticletitle \undefined \def \showarticletitle #1{#1}   \fi
\ifx \showURL      \undefined \def \showURL       {\relax}        \fi
\providecommand\bibfield[2]{#2}
\providecommand\bibinfo[2]{#2}
\providecommand\natexlab[1]{#1}
\providecommand\showeprint[2][]{arXiv:#2}

\bibitem[Bertiche et~al\mbox{.}(2020)]%
        {bertiche2020cloth3d}
\bibfield{author}{\bibinfo{person}{Hugo Bertiche}, \bibinfo{person}{Meysam
  Madadi}, {and} \bibinfo{person}{Sergio Escalera}.}
  \bibinfo{year}{2020}\natexlab{}.
\newblock \showarticletitle{CLOTH3D: clothed 3d humans}. In
  \bibinfo{booktitle}{\emph{Computer Vision--ECCV 2020: 16th European
  Conference, Glasgow, UK, August 23--28, 2020, Proceedings, Part XX 16}}.
  Springer, \bibinfo{pages}{344--359}.
\newblock


\bibitem[Chan et~al\mbox{.}(2022)]%
        {chan2022efficient}
\bibfield{author}{\bibinfo{person}{Eric~R Chan}, \bibinfo{person}{Connor~Z
  Lin}, \bibinfo{person}{Matthew~A Chan}, \bibinfo{person}{Koki Nagano},
  \bibinfo{person}{Boxiao Pan}, \bibinfo{person}{Shalini De~Mello},
  \bibinfo{person}{Orazio Gallo}, \bibinfo{person}{Leonidas~J Guibas},
  \bibinfo{person}{Jonathan Tremblay}, \bibinfo{person}{Sameh Khamis},
  {et~al\mbox{.}}} \bibinfo{year}{2022}\natexlab{}.
\newblock \showarticletitle{Efficient geometry-aware 3D generative adversarial
  networks}. In \bibinfo{booktitle}{\emph{Proceedings of the IEEE/CVF
  Conference on Computer Vision and Pattern Recognition}}.
  \bibinfo{pages}{16123--16133}.
\newblock


\bibitem[Cheng et~al\mbox{.}(2023)]%
        {samtrack}
\bibfield{author}{\bibinfo{person}{Yangming Cheng}, \bibinfo{person}{Liulei
  Li}, \bibinfo{person}{Yuanyou Xu}, \bibinfo{person}{Xiaodi Li},
  \bibinfo{person}{Zongxin Yang}, \bibinfo{person}{Wenguan Wang}, {and}
  \bibinfo{person}{Yi Yang}.} \bibinfo{year}{2023}\natexlab{}.
\newblock \showarticletitle{Segment and track anything}.
\newblock \bibinfo{journal}{\emph{arXiv preprint arXiv:2305.06558}}
  (\bibinfo{year}{2023}).
\newblock


\bibitem[Corona et~al\mbox{.}(2021)]%
        {corona2021smplicit}
\bibfield{author}{\bibinfo{person}{Enric Corona}, \bibinfo{person}{Albert
  Pumarola}, \bibinfo{person}{Guillem Alenya}, \bibinfo{person}{Gerard
  Pons-Moll}, {and} \bibinfo{person}{Francesc Moreno-Noguer}.}
  \bibinfo{year}{2021}\natexlab{}.
\newblock \showarticletitle{Smplicit: Topology-aware generative model for
  clothed people}. In \bibinfo{booktitle}{\emph{Proceedings of the IEEE/CVF
  conference on computer vision and pattern recognition}}.
  \bibinfo{pages}{11875--11885}.
\newblock


\bibitem[Ding et~al\mbox{.}(2022)]%
        {ding2022decoupling}
\bibfield{author}{\bibinfo{person}{Jian Ding}, \bibinfo{person}{Nan Xue},
  \bibinfo{person}{Gui-Song Xia}, {and} \bibinfo{person}{Dengxin Dai}.}
  \bibinfo{year}{2022}\natexlab{}.
\newblock \showarticletitle{Decoupling zero-shot semantic segmentation}. In
  \bibinfo{booktitle}{\emph{Proceedings of the IEEE/CVF Conference on Computer
  Vision and Pattern Recognition}}. \bibinfo{pages}{11583--11592}.
\newblock


\bibitem[Drebin et~al\mbox{.}(1988)]%
        {drebin1988volume}
\bibfield{author}{\bibinfo{person}{Robert~A Drebin}, \bibinfo{person}{Loren
  Carpenter}, {and} \bibinfo{person}{Pat Hanrahan}.}
  \bibinfo{year}{1988}\natexlab{}.
\newblock \showarticletitle{Volume rendering}.
\newblock \bibinfo{journal}{\emph{ACM Siggraph Computer Graphics}}
  \bibinfo{volume}{22}, \bibinfo{number}{4} (\bibinfo{year}{1988}),
  \bibinfo{pages}{65--74}.
\newblock


\bibitem[Feng et~al\mbox{.}(2022)]%
        {feng2022capturing}
\bibfield{author}{\bibinfo{person}{Yao Feng}, \bibinfo{person}{Jinlong Yang},
  \bibinfo{person}{Marc Pollefeys}, \bibinfo{person}{Michael~J Black}, {and}
  \bibinfo{person}{Timo Bolkart}.} \bibinfo{year}{2022}\natexlab{}.
\newblock \showarticletitle{Capturing and animation of body and clothing from
  monocular video}.
\newblock \bibinfo{journal}{\emph{arXiv preprint arXiv:2210.01868}}
  (\bibinfo{year}{2022}).
\newblock


\bibitem[Geitgey(2017)]%
        {geitgey2017face}
\bibfield{author}{\bibinfo{person}{Geitgey}.} \bibinfo{year}{2017}\natexlab{}.
\newblock \bibinfo{title}{Face Recognition}.
\newblock
\newblock
\urldef\tempurl%
\url{https://github.com/ageitgey/face_recognition}
\showURL{%
\tempurl}


\bibitem[Hong et~al\mbox{.}(2022)]%
        {hong2022avatarclip}
\bibfield{author}{\bibinfo{person}{Fangzhou Hong}, \bibinfo{person}{Mingyuan
  Zhang}, \bibinfo{person}{Liang Pan}, \bibinfo{person}{Zhongang Cai},
  \bibinfo{person}{Lei Yang}, {and} \bibinfo{person}{Ziwei Liu}.}
  \bibinfo{year}{2022}\natexlab{}.
\newblock \showarticletitle{Avatarclip: Zero-shot text-driven generation and
  animation of 3d avatars}.
\newblock \bibinfo{journal}{\emph{arXiv preprint arXiv:2205.08535}}
  (\bibinfo{year}{2022}).
\newblock


\bibitem[Jain et~al\mbox{.}(2022)]%
        {jain2022zero}
\bibfield{author}{\bibinfo{person}{Ajay Jain}, \bibinfo{person}{Ben
  Mildenhall}, \bibinfo{person}{Jonathan~T Barron}, \bibinfo{person}{Pieter
  Abbeel}, {and} \bibinfo{person}{Ben Poole}.} \bibinfo{year}{2022}\natexlab{}.
\newblock \showarticletitle{Zero-shot text-guided object generation with dream
  fields}. In \bibinfo{booktitle}{\emph{Proceedings of the IEEE/CVF Conference
  on Computer Vision and Pattern Recognition}}. \bibinfo{pages}{867--876}.
\newblock


\bibitem[Jiang et~al\mbox{.}(2020)]%
        {jiang2020bcnet}
\bibfield{author}{\bibinfo{person}{Boyi Jiang}, \bibinfo{person}{Juyong Zhang},
  \bibinfo{person}{Yang Hong}, \bibinfo{person}{Jinhao Luo},
  \bibinfo{person}{Ligang Liu}, {and} \bibinfo{person}{Hujun Bao}.}
  \bibinfo{year}{2020}\natexlab{}.
\newblock \showarticletitle{Bcnet: Learning body and cloth shape from a single
  image}. In \bibinfo{booktitle}{\emph{Computer Vision--ECCV 2020: 16th
  European Conference, Glasgow, UK, August 23--28, 2020, Proceedings, Part XX
  16}}. Springer, \bibinfo{pages}{18--35}.
\newblock


\bibitem[Jiang et~al\mbox{.}(2022)]%
        {jiang2022text2human}
\bibfield{author}{\bibinfo{person}{Yuming Jiang}, \bibinfo{person}{Shuai Yang},
  \bibinfo{person}{Haonan Qiu}, \bibinfo{person}{Wayne Wu},
  \bibinfo{person}{Chen~Change Loy}, {and} \bibinfo{person}{Ziwei Liu}.}
  \bibinfo{year}{2022}\natexlab{}.
\newblock \showarticletitle{Text2human: Text-driven controllable human image
  generation}.
\newblock \bibinfo{journal}{\emph{ACM Transactions on Graphics (TOG)}}
  \bibinfo{volume}{41}, \bibinfo{number}{4} (\bibinfo{year}{2022}),
  \bibinfo{pages}{1--11}.
\newblock


\bibitem[Lewis et~al\mbox{.}(2021)]%
        {lewis2021tryongan}
\bibfield{author}{\bibinfo{person}{Kathleen~M Lewis},
  \bibinfo{person}{Srivatsan Varadharajan}, {and} \bibinfo{person}{Ira
  Kemelmacher-Shlizerman}.} \bibinfo{year}{2021}\natexlab{}.
\newblock \showarticletitle{Tryongan: Body-aware try-on via layered
  interpolation}.
\newblock \bibinfo{journal}{\emph{ACM Transactions on Graphics (TOG)}}
  \bibinfo{volume}{40}, \bibinfo{number}{4} (\bibinfo{year}{2021}),
  \bibinfo{pages}{1--10}.
\newblock


\bibitem[Li et~al\mbox{.}(2018)]%
        {li2018differentiable}
\bibfield{author}{\bibinfo{person}{Tzu-Mao Li}, \bibinfo{person}{Miika
  Aittala}, \bibinfo{person}{Fr{\'e}do Durand}, {and} \bibinfo{person}{Jaakko
  Lehtinen}.} \bibinfo{year}{2018}\natexlab{}.
\newblock \showarticletitle{Differentiable monte carlo ray tracing through edge
  sampling}.
\newblock \bibinfo{journal}{\emph{ACM Transactions on Graphics (TOG)}}
  \bibinfo{volume}{37}, \bibinfo{number}{6} (\bibinfo{year}{2018}),
  \bibinfo{pages}{1--11}.
\newblock


\bibitem[Lin et~al\mbox{.}(2022)]%
        {lin2022magic3d}
\bibfield{author}{\bibinfo{person}{Chen-Hsuan Lin}, \bibinfo{person}{Jun Gao},
  \bibinfo{person}{Luming Tang}, \bibinfo{person}{Towaki Takikawa},
  \bibinfo{person}{Xiaohui Zeng}, \bibinfo{person}{Xun Huang},
  \bibinfo{person}{Karsten Kreis}, \bibinfo{person}{Sanja Fidler},
  \bibinfo{person}{Ming-Yu Liu}, {and} \bibinfo{person}{Tsung-Yi Lin}.}
  \bibinfo{year}{2022}\natexlab{}.
\newblock \showarticletitle{Magic3D: High-Resolution Text-to-3D Content
  Creation}.
\newblock \bibinfo{journal}{\emph{arXiv preprint arXiv:2211.10440}}
  (\bibinfo{year}{2022}).
\newblock


\bibitem[Loper et~al\mbox{.}(2015)]%
        {loper2015smpl}
\bibfield{author}{\bibinfo{person}{Matthew Loper}, \bibinfo{person}{Naureen
  Mahmood}, \bibinfo{person}{Javier Romero}, \bibinfo{person}{Gerard
  Pons-Moll}, {and} \bibinfo{person}{Michael~J Black}.}
  \bibinfo{year}{2015}\natexlab{}.
\newblock \showarticletitle{SMPL: A skinned multi-person linear model}.
\newblock \bibinfo{journal}{\emph{ACM transactions on graphics (TOG)}}
  \bibinfo{volume}{34}, \bibinfo{number}{6} (\bibinfo{year}{2015}),
  \bibinfo{pages}{1--16}.
\newblock


\bibitem[Luo and Hu(2021)]%
        {luo2021diffusion}
\bibfield{author}{\bibinfo{person}{Shitong Luo} {and} \bibinfo{person}{Wei
  Hu}.} \bibinfo{year}{2021}\natexlab{}.
\newblock \showarticletitle{Diffusion probabilistic models for 3d point cloud
  generation}. In \bibinfo{booktitle}{\emph{Proceedings of the IEEE/CVF
  Conference on Computer Vision and Pattern Recognition}}.
  \bibinfo{pages}{2837--2845}.
\newblock


\bibitem[Metzer et~al\mbox{.}(2022)]%
        {metzer2022latent}
\bibfield{author}{\bibinfo{person}{Gal Metzer}, \bibinfo{person}{Elad
  Richardson}, \bibinfo{person}{Or Patashnik}, \bibinfo{person}{Raja Giryes},
  {and} \bibinfo{person}{Daniel Cohen-Or}.} \bibinfo{year}{2022}\natexlab{}.
\newblock \showarticletitle{Latent-NeRF for Shape-Guided Generation of 3D
  Shapes and Textures}.
\newblock \bibinfo{journal}{\emph{arXiv preprint arXiv:2211.07600}}
  (\bibinfo{year}{2022}).
\newblock


\bibitem[Michel et~al\mbox{.}(2022)]%
        {michel2022text2mesh}
\bibfield{author}{\bibinfo{person}{Oscar Michel}, \bibinfo{person}{Roi Bar-On},
  \bibinfo{person}{Richard Liu}, \bibinfo{person}{Sagie Benaim}, {and}
  \bibinfo{person}{Rana Hanocka}.} \bibinfo{year}{2022}\natexlab{}.
\newblock \showarticletitle{Text2mesh: Text-driven neural stylization for
  meshes}. In \bibinfo{booktitle}{\emph{Proceedings of the IEEE/CVF Conference
  on Computer Vision and Pattern Recognition}}. \bibinfo{pages}{13492--13502}.
\newblock


\bibitem[Mildenhall et~al\mbox{.}(2021)]%
        {mildenhall2021nerf}
\bibfield{author}{\bibinfo{person}{Ben Mildenhall}, \bibinfo{person}{Pratul~P
  Srinivasan}, \bibinfo{person}{Matthew Tancik}, \bibinfo{person}{Jonathan~T
  Barron}, \bibinfo{person}{Ravi Ramamoorthi}, {and} \bibinfo{person}{Ren Ng}.}
  \bibinfo{year}{2021}\natexlab{}.
\newblock \showarticletitle{Nerf: Representing scenes as neural radiance fields
  for view synthesis}.
\newblock \bibinfo{journal}{\emph{Commun. ACM}} \bibinfo{volume}{65},
  \bibinfo{number}{1} (\bibinfo{year}{2021}), \bibinfo{pages}{99--106}.
\newblock


\bibitem[Mokady et~al\mbox{.}(2021)]%
        {mokady2021clipcap}
\bibfield{author}{\bibinfo{person}{Ron Mokady}, \bibinfo{person}{Amir Hertz},
  {and} \bibinfo{person}{Amit~H Bermano}.} \bibinfo{year}{2021}\natexlab{}.
\newblock \showarticletitle{Clipcap: Clip prefix for image captioning}.
\newblock \bibinfo{journal}{\emph{arXiv preprint arXiv:2111.09734}}
  (\bibinfo{year}{2021}).
\newblock


\bibitem[M{\"u}ller et~al\mbox{.}(2019)]%
        {muller2019neural}
\bibfield{author}{\bibinfo{person}{Thomas M{\"u}ller}, \bibinfo{person}{Brian
  McWilliams}, \bibinfo{person}{Fabrice Rousselle}, \bibinfo{person}{Markus
  Gross}, {and} \bibinfo{person}{Jan Nov{\'a}k}.}
  \bibinfo{year}{2019}\natexlab{}.
\newblock \showarticletitle{Neural importance sampling}.
\newblock \bibinfo{journal}{\emph{ACM Transactions on Graphics (ToG)}}
  \bibinfo{volume}{38}, \bibinfo{number}{5} (\bibinfo{year}{2019}),
  \bibinfo{pages}{1--19}.
\newblock


\bibitem[Nichol et~al\mbox{.}(2021)]%
        {nichol2021glide}
\bibfield{author}{\bibinfo{person}{Alex Nichol}, \bibinfo{person}{Prafulla
  Dhariwal}, \bibinfo{person}{Aditya Ramesh}, \bibinfo{person}{Pranav Shyam},
  \bibinfo{person}{Pamela Mishkin}, \bibinfo{person}{Bob McGrew},
  \bibinfo{person}{Ilya Sutskever}, {and} \bibinfo{person}{Mark Chen}.}
  \bibinfo{year}{2021}\natexlab{}.
\newblock \showarticletitle{Glide: Towards photorealistic image generation and
  editing with text-guided diffusion models}.
\newblock \bibinfo{journal}{\emph{arXiv preprint arXiv:2112.10741}}
  (\bibinfo{year}{2021}).
\newblock


\bibitem[Nichol et~al\mbox{.}(2022)]%
        {nichol2022point}
\bibfield{author}{\bibinfo{person}{Alex Nichol}, \bibinfo{person}{Heewoo Jun},
  \bibinfo{person}{Prafulla Dhariwal}, \bibinfo{person}{Pamela Mishkin}, {and}
  \bibinfo{person}{Mark Chen}.} \bibinfo{year}{2022}\natexlab{}.
\newblock \showarticletitle{Point-E: A System for Generating 3D Point Clouds
  from Complex Prompts}.
\newblock \bibinfo{journal}{\emph{arXiv preprint arXiv:2212.08751}}
  (\bibinfo{year}{2022}).
\newblock


\bibitem[Niemeyer et~al\mbox{.}(2020)]%
        {niemeyer2020differentiable}
\bibfield{author}{\bibinfo{person}{Michael Niemeyer}, \bibinfo{person}{Lars
  Mescheder}, \bibinfo{person}{Michael Oechsle}, {and} \bibinfo{person}{Andreas
  Geiger}.} \bibinfo{year}{2020}\natexlab{}.
\newblock \showarticletitle{Differentiable volumetric rendering: Learning
  implicit 3d representations without 3d supervision}. In
  \bibinfo{booktitle}{\emph{Proceedings of the IEEE/CVF Conference on Computer
  Vision and Pattern Recognition}}. \bibinfo{pages}{3504--3515}.
\newblock


\bibitem[Or-El et~al\mbox{.}(2022)]%
        {or2022stylesdf}
\bibfield{author}{\bibinfo{person}{Roy Or-El}, \bibinfo{person}{Xuan Luo},
  \bibinfo{person}{Mengyi Shan}, \bibinfo{person}{Eli Shechtman},
  \bibinfo{person}{Jeong~Joon Park}, {and} \bibinfo{person}{Ira
  Kemelmacher-Shlizerman}.} \bibinfo{year}{2022}\natexlab{}.
\newblock \showarticletitle{Stylesdf: High-resolution 3d-consistent image and
  geometry generation}. In \bibinfo{booktitle}{\emph{Proceedings of the
  IEEE/CVF Conference on Computer Vision and Pattern Recognition}}.
  \bibinfo{pages}{13503--13513}.
\newblock


\bibitem[Pan et~al\mbox{.}(2023)]%
        {pan2023transhuman}
\bibfield{author}{\bibinfo{person}{Xiao Pan}, \bibinfo{person}{Zongxin Yang},
  \bibinfo{person}{Jianxin Ma}, \bibinfo{person}{Chang Zhou}, {and}
  \bibinfo{person}{Yi Yang}.} \bibinfo{year}{2023}\natexlab{}.
\newblock \showarticletitle{TransHuman: A Transformer-based Human
  Representation for Generalizable Neural Human Rendering}.
\newblock \bibinfo{journal}{\emph{Proceedings of the IEEE/CVF International
  conference on computer vision}}.
\newblock


\bibitem[Park et~al\mbox{.}(2021)]%
        {park2021nerfies}
\bibfield{author}{\bibinfo{person}{Keunhong Park}, \bibinfo{person}{Utkarsh
  Sinha}, \bibinfo{person}{Jonathan~T Barron}, \bibinfo{person}{Sofien
  Bouaziz}, \bibinfo{person}{Dan~B Goldman}, \bibinfo{person}{Steven~M Seitz},
  {and} \bibinfo{person}{Ricardo Martin-Brualla}.}
  \bibinfo{year}{2021}\natexlab{}.
\newblock \showarticletitle{Nerfies: Deformable neural radiance fields}. In
  \bibinfo{booktitle}{\emph{Proceedings of the IEEE/CVF International
  Conference on Computer Vision}}. \bibinfo{pages}{5865--5874}.
\newblock


\bibitem[Park et~al\mbox{.}(2019)]%
        {park2019semantic}
\bibfield{author}{\bibinfo{person}{Taesung Park}, \bibinfo{person}{Ming-Yu
  Liu}, \bibinfo{person}{Ting-Chun Wang}, {and} \bibinfo{person}{Jun-Yan Zhu}.}
  \bibinfo{year}{2019}\natexlab{}.
\newblock \showarticletitle{Semantic image synthesis with spatially-adaptive
  normalization}. In \bibinfo{booktitle}{\emph{Proceedings of the IEEE/CVF
  conference on computer vision and pattern recognition}}.
  \bibinfo{pages}{2337--2346}.
\newblock


\bibitem[Patel et~al\mbox{.}(2020)]%
        {patel2020tailornet}
\bibfield{author}{\bibinfo{person}{Chaitanya Patel},
  \bibinfo{person}{Zhouyingcheng Liao}, {and} \bibinfo{person}{Gerard
  Pons-Moll}.} \bibinfo{year}{2020}\natexlab{}.
\newblock \showarticletitle{Tailornet: Predicting clothing in 3d as a function
  of human pose, shape and garment style}. In
  \bibinfo{booktitle}{\emph{Proceedings of the IEEE/CVF Conference on Computer
  Vision and Pattern Recognition}}. \bibinfo{pages}{7365--7375}.
\newblock


\bibitem[Peng et~al\mbox{.}(2021)]%
        {peng2021animatable}
\bibfield{author}{\bibinfo{person}{Sida Peng}, \bibinfo{person}{Junting Dong},
  \bibinfo{person}{Qianqian Wang}, \bibinfo{person}{Shangzhan Zhang},
  \bibinfo{person}{Qing Shuai}, \bibinfo{person}{Xiaowei Zhou}, {and}
  \bibinfo{person}{Hujun Bao}.} \bibinfo{year}{2021}\natexlab{}.
\newblock \showarticletitle{Animatable neural radiance fields for modeling
  dynamic human bodies}. In \bibinfo{booktitle}{\emph{Proceedings of the
  IEEE/CVF International Conference on Computer Vision}}.
  \bibinfo{pages}{14314--14323}.
\newblock


\bibitem[Poole et~al\mbox{.}(2022)]%
        {poole2022dreamfusion}
\bibfield{author}{\bibinfo{person}{Ben Poole}, \bibinfo{person}{Ajay Jain},
  \bibinfo{person}{Jonathan~T Barron}, {and} \bibinfo{person}{Ben Mildenhall}.}
  \bibinfo{year}{2022}\natexlab{}.
\newblock \showarticletitle{Dreamfusion: Text-to-3d using 2d diffusion}.
\newblock \bibinfo{journal}{\emph{arXiv preprint arXiv:2209.14988}}
  (\bibinfo{year}{2022}).
\newblock


\bibitem[Pumarola et~al\mbox{.}(2021)]%
        {pumarola2021d}
\bibfield{author}{\bibinfo{person}{Albert Pumarola}, \bibinfo{person}{Enric
  Corona}, \bibinfo{person}{Gerard Pons-Moll}, {and} \bibinfo{person}{Francesc
  Moreno-Noguer}.} \bibinfo{year}{2021}\natexlab{}.
\newblock \showarticletitle{D-nerf: Neural radiance fields for dynamic scenes}.
  In \bibinfo{booktitle}{\emph{Proceedings of the IEEE/CVF Conference on
  Computer Vision and Pattern Recognition}}. \bibinfo{pages}{10318--10327}.
\newblock


\bibitem[Radford et~al\mbox{.}(2021)]%
        {radford2021learning}
\bibfield{author}{\bibinfo{person}{Alec Radford}, \bibinfo{person}{Jong~Wook
  Kim}, \bibinfo{person}{Chris Hallacy}, \bibinfo{person}{Aditya Ramesh},
  \bibinfo{person}{Gabriel Goh}, \bibinfo{person}{Sandhini Agarwal},
  \bibinfo{person}{Girish Sastry}, \bibinfo{person}{Amanda Askell},
  \bibinfo{person}{Pamela Mishkin}, \bibinfo{person}{Jack Clark},
  {et~al\mbox{.}}} \bibinfo{year}{2021}\natexlab{}.
\newblock \showarticletitle{Learning transferable visual models from natural
  language supervision}. In \bibinfo{booktitle}{\emph{International conference
  on machine learning}}. PMLR, \bibinfo{pages}{8748--8763}.
\newblock


\bibitem[Ramesh et~al\mbox{.}(2022)]%
        {ramesh2022hierarchical}
\bibfield{author}{\bibinfo{person}{Aditya Ramesh}, \bibinfo{person}{Prafulla
  Dhariwal}, \bibinfo{person}{Alex Nichol}, \bibinfo{person}{Casey Chu}, {and}
  \bibinfo{person}{Mark Chen}.} \bibinfo{year}{2022}\natexlab{}.
\newblock \showarticletitle{Hierarchical text-conditional image generation with
  clip latents}.
\newblock \bibinfo{journal}{\emph{arXiv preprint arXiv:2204.06125}}
  (\bibinfo{year}{2022}).
\newblock


\bibitem[Richardson et~al\mbox{.}(2023)]%
        {richardson2023texture}
\bibfield{author}{\bibinfo{person}{Elad Richardson}, \bibinfo{person}{Gal
  Metzer}, \bibinfo{person}{Yuval Alaluf}, \bibinfo{person}{Raja Giryes}, {and}
  \bibinfo{person}{Daniel Cohen-Or}.} \bibinfo{year}{2023}\natexlab{}.
\newblock \showarticletitle{Texture: Text-guided texturing of 3d shapes}.
\newblock \bibinfo{journal}{\emph{arXiv preprint arXiv:2302.01721}}
  (\bibinfo{year}{2023}).
\newblock


\bibitem[Rombach et~al\mbox{.}(2022)]%
        {rombach2022high}
\bibfield{author}{\bibinfo{person}{Robin Rombach}, \bibinfo{person}{Andreas
  Blattmann}, \bibinfo{person}{Dominik Lorenz}, \bibinfo{person}{Patrick
  Esser}, {and} \bibinfo{person}{Bj{\"o}rn Ommer}.}
  \bibinfo{year}{2022}\natexlab{}.
\newblock \showarticletitle{High-resolution image synthesis with latent
  diffusion models}. In \bibinfo{booktitle}{\emph{Proceedings of the IEEE/CVF
  Conference on Computer Vision and Pattern Recognition}}.
  \bibinfo{pages}{10684--10695}.
\newblock


\bibitem[Saharia et~al\mbox{.}(2022)]%
        {saharia2022photorealistic}
\bibfield{author}{\bibinfo{person}{Chitwan Saharia}, \bibinfo{person}{William
  Chan}, \bibinfo{person}{Saurabh Saxena}, \bibinfo{person}{Lala Li},
  \bibinfo{person}{Jay Whang}, \bibinfo{person}{Emily~L Denton},
  \bibinfo{person}{Kamyar Ghasemipour}, \bibinfo{person}{Raphael
  Gontijo~Lopes}, \bibinfo{person}{Burcu Karagol~Ayan}, \bibinfo{person}{Tim
  Salimans}, {et~al\mbox{.}}} \bibinfo{year}{2022}\natexlab{}.
\newblock \showarticletitle{Photorealistic text-to-image diffusion models with
  deep language understanding}.
\newblock \bibinfo{journal}{\emph{Advances in Neural Information Processing
  Systems}}  \bibinfo{volume}{35} (\bibinfo{year}{2022}),
  \bibinfo{pages}{36479--36494}.
\newblock


\bibitem[Santesteban et~al\mbox{.}(2019)]%
        {santesteban2019learning}
\bibfield{author}{\bibinfo{person}{Igor Santesteban}, \bibinfo{person}{Miguel~A
  Otaduy}, {and} \bibinfo{person}{Dan Casas}.} \bibinfo{year}{2019}\natexlab{}.
\newblock \showarticletitle{Learning-based animation of clothing for virtual
  try-on}. In \bibinfo{booktitle}{\emph{Computer Graphics Forum}},
  Vol.~\bibinfo{volume}{38}. Wiley Online Library, \bibinfo{pages}{355--366}.
\newblock


\bibitem[Seitzer(2020)]%
        {Seitzer2020FID}
\bibfield{author}{\bibinfo{person}{Maximilian Seitzer}.}
  \bibinfo{year}{2020}\natexlab{}.
\newblock \bibinfo{title}{{pytorch-fid: FID Score for PyTorch}}.
\newblock
  \bibinfo{howpublished}{\url{https://github.com/mseitzer/pytorch-fid}}.
\newblock
\newblock
\shownote{Version 0.3.0}.


\bibitem[Tang(2022)]%
        {stable-dreamfusion}
\bibfield{author}{\bibinfo{person}{Jiaxiang Tang}.}
  \bibinfo{year}{2022}\natexlab{}.
\newblock \bibinfo{title}{Stable-dreamfusion: Text-to-3D with
  Stable-diffusion}.
\newblock
\newblock
\newblock
\shownote{https://github.com/ashawkey/stable-dreamfusion}.


\bibitem[Tewel et~al\mbox{.}(2021)]%
        {tewel2021zero}
\bibfield{author}{\bibinfo{person}{Yoad Tewel}, \bibinfo{person}{Yoav Shalev},
  \bibinfo{person}{Idan Schwartz}, {and} \bibinfo{person}{Lior Wolf}.}
  \bibinfo{year}{2021}\natexlab{}.
\newblock \showarticletitle{Zero-shot image-to-text generation for
  visual-semantic arithmetic}.
\newblock \bibinfo{journal}{\emph{arXiv preprint arXiv:2111.14447}}
  (\bibinfo{year}{2021}).
\newblock


\bibitem[Vidaurre et~al\mbox{.}(2020)]%
        {vidaurre2020fully}
\bibfield{author}{\bibinfo{person}{Raquel Vidaurre}, \bibinfo{person}{Igor
  Santesteban}, \bibinfo{person}{Elena Garces}, {and} \bibinfo{person}{Dan
  Casas}.} \bibinfo{year}{2020}\natexlab{}.
\newblock \showarticletitle{Fully convolutional graph neural networks for
  parametric virtual try-on}. In \bibinfo{booktitle}{\emph{Computer Graphics
  Forum}}, Vol.~\bibinfo{volume}{39}. Wiley Online Library,
  \bibinfo{pages}{145--156}.
\newblock


\bibitem[Wang et~al\mbox{.}(2022a)]%
        {wang2022nerf}
\bibfield{author}{\bibinfo{person}{Can Wang}, \bibinfo{person}{Ruixiang Jiang},
  \bibinfo{person}{Menglei Chai}, \bibinfo{person}{Mingming He},
  \bibinfo{person}{Dongdong Chen}, {and} \bibinfo{person}{Jing Liao}.}
  \bibinfo{year}{2022}\natexlab{a}.
\newblock \showarticletitle{NeRF-Art: Text-Driven Neural Radiance Fields
  Stylization}.
\newblock \bibinfo{journal}{\emph{arXiv preprint arXiv:2212.08070}}
  (\bibinfo{year}{2022}).
\newblock


\bibitem[Wang et~al\mbox{.}(2021)]%
        {wang2021neus}
\bibfield{author}{\bibinfo{person}{Peng Wang}, \bibinfo{person}{Lingjie Liu},
  \bibinfo{person}{Yuan Liu}, \bibinfo{person}{Christian Theobalt},
  \bibinfo{person}{Taku Komura}, {and} \bibinfo{person}{Wenping Wang}.}
  \bibinfo{year}{2021}\natexlab{}.
\newblock \showarticletitle{Neus: Learning neural implicit surfaces by volume
  rendering for multi-view reconstruction}.
\newblock \bibinfo{journal}{\emph{arXiv preprint arXiv:2106.10689}}
  (\bibinfo{year}{2021}).
\newblock


\bibitem[Wang et~al\mbox{.}(2022c)]%
        {wang2022rodin}
\bibfield{author}{\bibinfo{person}{Tengfei Wang}, \bibinfo{person}{Bo Zhang},
  \bibinfo{person}{Ting Zhang}, \bibinfo{person}{Shuyang Gu},
  \bibinfo{person}{Jianmin Bao}, \bibinfo{person}{Tadas Baltrusaitis},
  \bibinfo{person}{Jingjing Shen}, \bibinfo{person}{Dong Chen},
  \bibinfo{person}{Fang Wen}, \bibinfo{person}{Qifeng Chen}, {et~al\mbox{.}}}
  \bibinfo{year}{2022}\natexlab{c}.
\newblock \showarticletitle{Rodin: A Generative Model for Sculpting 3D Digital
  Avatars Using Diffusion}.
\newblock \bibinfo{journal}{\emph{arXiv preprint arXiv:2212.06135}}
  (\bibinfo{year}{2022}).
\newblock


\bibitem[Wang et~al\mbox{.}(2018)]%
        {wang2018high}
\bibfield{author}{\bibinfo{person}{Ting-Chun Wang}, \bibinfo{person}{Ming-Yu
  Liu}, \bibinfo{person}{Jun-Yan Zhu}, \bibinfo{person}{Andrew Tao},
  \bibinfo{person}{Jan Kautz}, {and} \bibinfo{person}{Bryan Catanzaro}.}
  \bibinfo{year}{2018}\natexlab{}.
\newblock \showarticletitle{High-resolution image synthesis and semantic
  manipulation with conditional gans}. In \bibinfo{booktitle}{\emph{Proceedings
  of the IEEE conference on computer vision and pattern recognition}}.
  \bibinfo{pages}{8798--8807}.
\newblock


\bibitem[Wang et~al\mbox{.}(2022b)]%
        {wang2022cris}
\bibfield{author}{\bibinfo{person}{Zhaoqing Wang}, \bibinfo{person}{Yu Lu},
  \bibinfo{person}{Qiang Li}, \bibinfo{person}{Xunqiang Tao},
  \bibinfo{person}{Yandong Guo}, \bibinfo{person}{Mingming Gong}, {and}
  \bibinfo{person}{Tongliang Liu}.} \bibinfo{year}{2022}\natexlab{b}.
\newblock \showarticletitle{Cris: Clip-driven referring image segmentation}. In
  \bibinfo{booktitle}{\emph{Proceedings of the IEEE/CVF conference on computer
  vision and pattern recognition}}. \bibinfo{pages}{11686--11695}.
\newblock


\bibitem[Weng et~al\mbox{.}(2020)]%
        {weng2020misc}
\bibfield{author}{\bibinfo{person}{Shuchen Weng}, \bibinfo{person}{Wenbo Li},
  \bibinfo{person}{Dawei Li}, \bibinfo{person}{Hongxia Jin}, {and}
  \bibinfo{person}{Boxin Shi}.} \bibinfo{year}{2020}\natexlab{}.
\newblock \showarticletitle{Misc: Multi-condition injection and
  spatially-adaptive compositing for conditional person image synthesis}. In
  \bibinfo{booktitle}{\emph{Proceedings of the IEEE/CVF Conference on Computer
  Vision and Pattern Recognition}}. \bibinfo{pages}{7741--7749}.
\newblock


\bibitem[Xiang et~al\mbox{.}(2021)]%
        {xiang2021modeling}
\bibfield{author}{\bibinfo{person}{Donglai Xiang}, \bibinfo{person}{Fabian
  Prada}, \bibinfo{person}{Timur Bagautdinov}, \bibinfo{person}{Weipeng Xu},
  \bibinfo{person}{Yuan Dong}, \bibinfo{person}{He Wen},
  \bibinfo{person}{Jessica Hodgins}, {and} \bibinfo{person}{Chenglei Wu}.}
  \bibinfo{year}{2021}\natexlab{}.
\newblock \showarticletitle{Modeling clothing as a separate layer for an
  animatable human avatar}.
\newblock \bibinfo{journal}{\emph{ACM Transactions on Graphics (TOG)}}
  \bibinfo{volume}{40}, \bibinfo{number}{6} (\bibinfo{year}{2021}),
  \bibinfo{pages}{1--15}.
\newblock


\bibitem[Xu et~al\mbox{.}(2021)]%
        {xu2021simple}
\bibfield{author}{\bibinfo{person}{Mengde Xu}, \bibinfo{person}{Zheng Zhang},
  \bibinfo{person}{Fangyun Wei}, \bibinfo{person}{Yutong Lin},
  \bibinfo{person}{Yue Cao}, \bibinfo{person}{Han Hu}, {and}
  \bibinfo{person}{Xiang Bai}.} \bibinfo{year}{2021}\natexlab{}.
\newblock \showarticletitle{A simple baseline for zero-shot semantic
  segmentation with pre-trained vision-language model}.
\newblock \bibinfo{journal}{\emph{arXiv preprint arXiv:2112.14757}}
  (\bibinfo{year}{2021}).
\newblock


\bibitem[Yang et~al\mbox{.}(2021)]%
        {mkr}
\bibfield{author}{\bibinfo{person}{Yi Yang}, \bibinfo{person}{Yueting Zhuang},
  {and} \bibinfo{person}{Yunhe Pan}.} \bibinfo{year}{2021}\natexlab{}.
\newblock \showarticletitle{Multiple knowledge representation for big data
  artificial intelligence: framework, applications, and case studies}.
\newblock \bibinfo{journal}{\emph{Frontiers of Information Technology \&
  Electronic Engineering}} \bibinfo{volume}{22}, \bibinfo{number}{12}
  (\bibinfo{year}{2021}), \bibinfo{pages}{1551--1558}.
\newblock


\bibitem[Yoon et~al\mbox{.}(2021)]%
        {yoon2021pose}
\bibfield{author}{\bibinfo{person}{Jae~Shin Yoon}, \bibinfo{person}{Lingjie
  Liu}, \bibinfo{person}{Vladislav Golyanik}, \bibinfo{person}{Kripasindhu
  Sarkar}, \bibinfo{person}{Hyun~Soo Park}, {and} \bibinfo{person}{Christian
  Theobalt}.} \bibinfo{year}{2021}\natexlab{}.
\newblock \showarticletitle{Pose-guided human animation from a single image in
  the wild}. In \bibinfo{booktitle}{\emph{Proceedings of the IEEE/CVF
  Conference on Computer Vision and Pattern Recognition}}.
  \bibinfo{pages}{15039--15048}.
\newblock


\bibitem[Zhang et~al\mbox{.}(2023)]%
        {zhang2023avatargen}
\bibfield{author}{\bibinfo{person}{Jianfeng Zhang}, \bibinfo{person}{Zihang
  Jiang}, \bibinfo{person}{Dingdong Yang}, \bibinfo{person}{Hongyi Xu},
  \bibinfo{person}{Yichun Shi}, \bibinfo{person}{Guoxian Song},
  \bibinfo{person}{Zhongcong Xu}, \bibinfo{person}{Xinchao Wang}, {and}
  \bibinfo{person}{Jiashi Feng}.} \bibinfo{year}{2023}\natexlab{}.
\newblock \showarticletitle{Avatargen: a 3d generative model for animatable
  human avatars}. In \bibinfo{booktitle}{\emph{Computer Vision--ECCV 2022
  Workshops: Tel Aviv, Israel, October 23--27, 2022, Proceedings, Part III}}.
  Springer, \bibinfo{pages}{668--685}.
\newblock


\bibitem[Zhou et~al\mbox{.}(2021)]%
        {zhou20213d}
\bibfield{author}{\bibinfo{person}{Linqi Zhou}, \bibinfo{person}{Yilun Du},
  {and} \bibinfo{person}{Jiajun Wu}.} \bibinfo{year}{2021}\natexlab{}.
\newblock \showarticletitle{3d shape generation and completion through
  point-voxel diffusion}. In \bibinfo{booktitle}{\emph{Proceedings of the
  IEEE/CVF International Conference on Computer Vision}}.
  \bibinfo{pages}{5826--5835}.
\newblock


\bibitem[Zhu et~al\mbox{.}(2020)]%
        {zhu2020deep}
\bibfield{author}{\bibinfo{person}{Heming Zhu}, \bibinfo{person}{Yu Cao},
  \bibinfo{person}{Hang Jin}, \bibinfo{person}{Weikai Chen},
  \bibinfo{person}{Dong Du}, \bibinfo{person}{Zhangye Wang},
  \bibinfo{person}{Shuguang Cui}, {and} \bibinfo{person}{Xiaoguang Han}.}
  \bibinfo{year}{2020}\natexlab{}.
\newblock \showarticletitle{Deep fashion3d: A dataset and benchmark for 3d
  garment reconstruction from single images}. In
  \bibinfo{booktitle}{\emph{Computer Vision--ECCV 2020: 16th European
  Conference, Glasgow, UK, August 23--28, 2020, Proceedings, Part I 16}}.
  Springer, \bibinfo{pages}{512--530}.
\newblock


\end{thebibliography}
\clearpage
\appendix
\section{Supplementary Material}
\begin{figure*}[htbp]
\centering
\begin{minipage}{0.45\linewidth}
    \includegraphics[width=\linewidth]{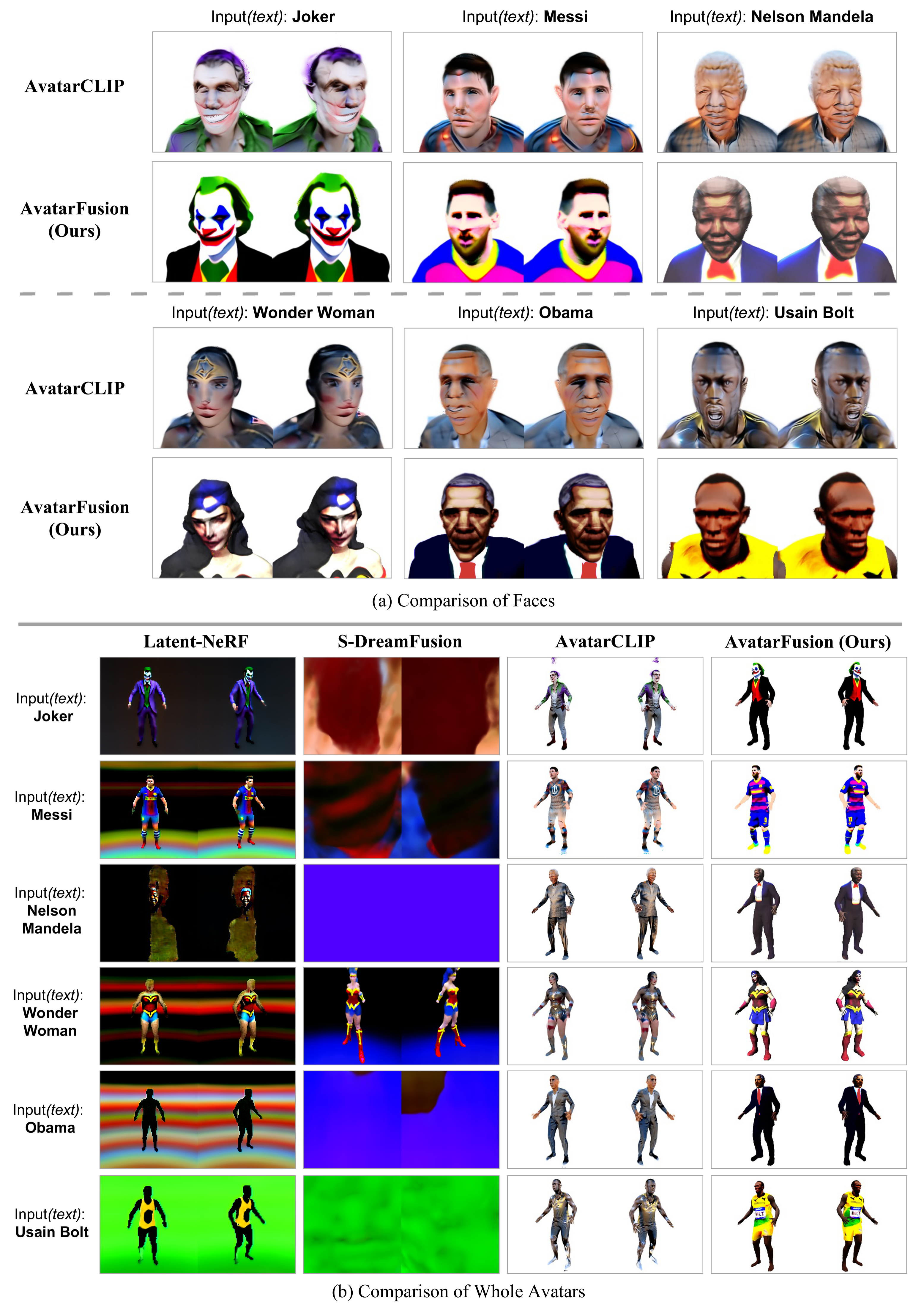}
    \caption{Comparison with Baselines.}
    \label{fig:baseline_more}
\end{minipage}
\begin{minipage}{0.49\linewidth}
    \includegraphics[width=\linewidth]{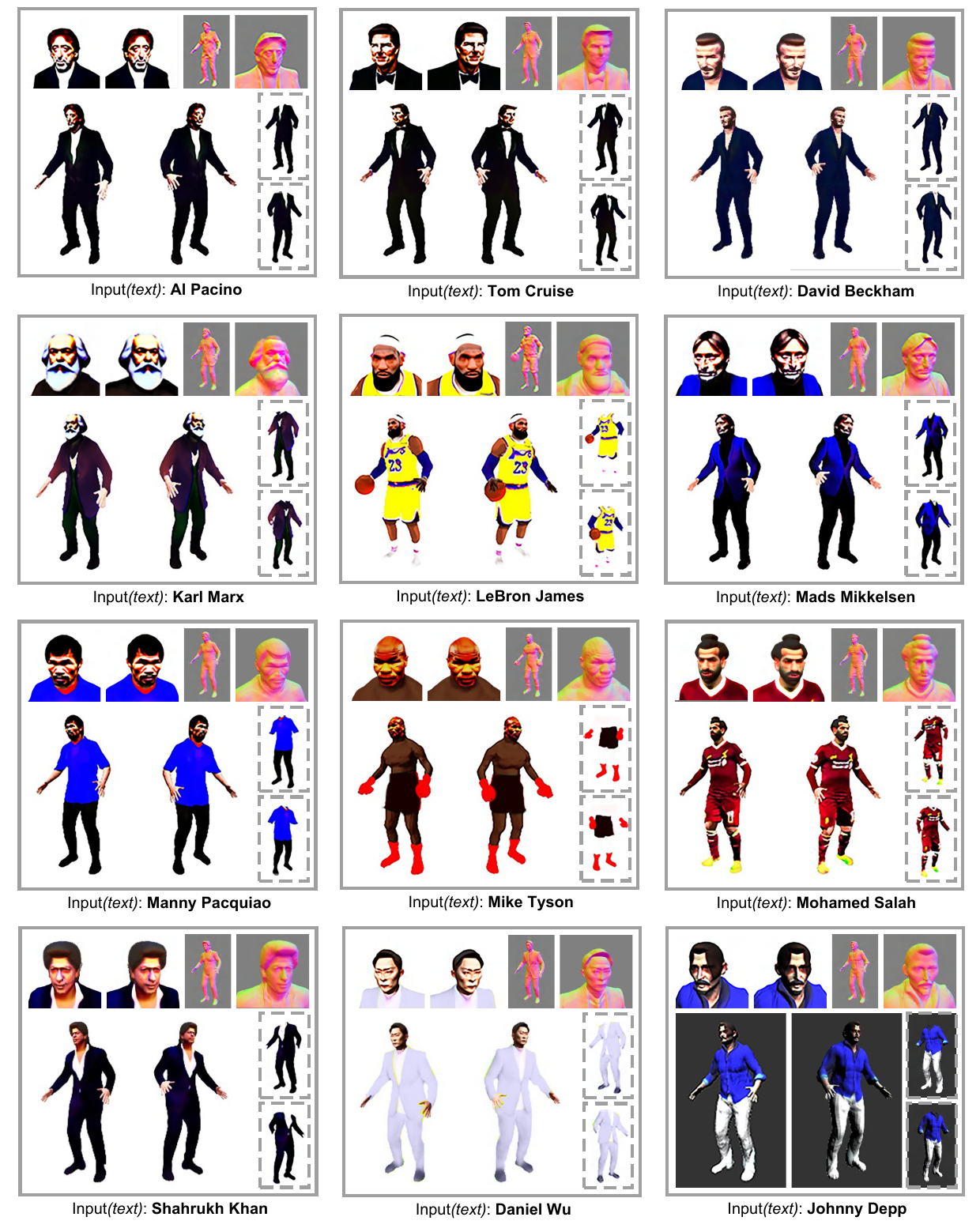}
    \caption{More Results of AvatarFusion.}
    \label{fig:results_more}
\end{minipage}
\end{figure*}
\subsection{Details of Methodology}
\subsubsection{Details of Dual Volume Rendering}
Section 4.3 presents the equations used to merge the skin and clothing spaces in the clothing-decoupled model. In this section, we provide the derivation of these equations.

Initially, to compute the color for each pixel, we shoot a ray $\mathbf{p}(t)$ and sample points on it, denoted as $\{\mathbf{x}_0, \mathbf{x}_1, \ldots, \mathbf{x}_i, \ldots\}$, where $\mathbf{x}_i=\mathbf{p}(t_{i})$. Next, we compute the SDF values of each point from the skin and clothing models, represented as $f_\mathrm{body}(\mathbf{x}_i)$, $f_\mathrm{cloth}(\mathbf{x}_i)$, and the corresponding RGB color vectors $\mathbf{c}_\mathrm{body}(\mathbf{x}_i)$, $\mathbf{c}_\mathrm{cloth}(\mathbf{x}_i)$, respectively.
Subsequently, we compute the discrete opacity values based on the SDF-based volume rendering technique \cite{wang2021neus}, which can be expressed as:

\begin{equation}
    \alpha_{\mathrm{body}}(\mathbf{x}_i) = \mathrm{max} \{ \frac{\Phi_{s}(f_\mathrm{body}(\mathbf{x}_i))- \Phi_{s}(f_\mathrm{body}(\mathbf{x}_{i+1}))}{\Phi_{s}(f_\mathrm{body}(\mathbf{x}_i))}, 0 \},
\end{equation}
\begin{equation}
    \alpha_{\mathrm{cloth}}(\mathbf{x}_i) = \mathrm{max} \{ \frac{\Phi_{s}(f_\mathrm{cloth}(\mathbf{x}_i))- \Phi_{s}(f_\mathrm{cloth}(\mathbf{x}_{i+1}))}{\Phi_{s}(f_\mathrm{cloth}(\mathbf{x}_i))}, 0 \},
\end{equation}
where $\Phi_{s}(x) = (1+e^{-sx})^{-1}$. This equation is a discretization of $\alpha_{i}=1-\mathrm{exp}(-\int^{t_{i+1}}_{t_{i}} {\rho(t)\mathrm{d}t}).$

For the inner space, where only skin space points are present, we use the characteristics obtained from the skin model as the properties of the joint space points. This can be denoted as:

\begin{equation}
    \alpha_i = \alpha_{\mathrm{body}}(\mathbf{x}_i),
   \quad f_{\mathrm{body}}(\mathbf{x}_i) <= \delta,
\end{equation}

\begin{equation}
    \mathbf{c}_i = \mathbf{c}_{\mathrm{body}}(\mathbf{x}_i),
    \quad f_{\mathrm{body}}(\mathbf{x}_i) <= \delta,
\end{equation}
where the inner space refers to the space where the distance from the skin is less than $\delta$.

For the outer space, where both skin space points and clothing space points exist, we need to merge the points from both spaces. To accomplish this, we begin by exploring the equations of volume rendering before discretization.
It is worth noting that the opaque density $\rho$ is proportional to the gas density in the primitive physical equations of volume rendering, as reported in \cite{drebin1988volume}. Based on this relationship, we can approximate that the volume density $\rho$ of the volume rendering functions can be added in the same way as gas densities are added. This can be expressed as:

\begin{equation}
\rho(t) = \rho_{\mathrm{body}}(t) + \rho_{\mathrm{cloth}}(t),
\end{equation}
where $\rho_{\mathrm{body}}(t)$ and $\rho_{\mathrm{cloth}}(t)$ represent the densities of the skin and clothing spaces, respectively, at time $t$ on the emitted ray.
Then, the discrete opacity value should be calculated as follow:
\begin{equation}
\begin{aligned}
    \alpha_{i}
    = & 1-\mathrm{exp}(-\int^{t_{i+1}}_{t_{i}}\rho(t)\mathrm{d}t)\\
    = & 1-\mathrm{exp}(-\int^{t_{i+1}}_{t_{i}} {(\rho_{\mathrm{body}}(t) + \rho_{\mathrm{cloth}}(t))\mathrm{d}t}) \\
    = & (1 - \mathrm{exp}(-\int^{t_{i+1}}_{t_{i}}{\rho_{\mathrm{body}}(t)\mathrm{d}t}))  + (1 - \mathrm{exp}(-\int^{t_{i+1}}_{t_{i}}{\rho_{\mathrm{cloth}}(t)\mathrm{d}t})) \\ & - (1 - \mathrm{exp}(-\int^{t_{i+1}}_{t_{i}}{\rho_{\mathrm{body}}(t)\mathrm{d}t})) \cdot (1 - \mathrm{exp}(-\int^{t_{i+1}}_{t_{i}}{\rho_{\mathrm{cloth}}(t)\mathrm{d}t})) \\
    = & \alpha_{\mathrm{body}}(\mathbf{x}_i) + \alpha_{\mathrm{cloth}}(\mathbf{x}_i) - \alpha_{\mathrm{body}}(\mathbf{x}_i) \cdot \alpha_{\mathrm{cloth}}(\mathbf{x}_i), \quad f_{\mathrm{body}}(\mathbf{x}_i) > \delta.
\end{aligned}
\end{equation}

For the color item $\mathbf{c}_{i}$, because there is no physical meaning of adding the RGB values, we make a simple approximation that 
\begin{equation}
\begin{split}
    \mathbf{c}_{i}=\frac{\alpha_{\mathrm{body}}(\mathbf{x}_i)}{\alpha_{\mathrm{body}}(\mathbf{x}_i) + \alpha_{\mathrm{cloth}}(\mathbf{x}_i)}\mathbf{c}_{\mathrm{body}}(\mathbf{x}_i)+\frac{\alpha_{\mathrm{cloth}}(\mathbf{x}_i)}{\alpha_{\mathrm{body}}(\mathbf{x}_i) + \alpha_{\mathrm{cloth}}(\mathbf{x}_i)} \\
    \cdot \mathbf{c}_{\mathrm{cloth}}(\mathbf{x}_i), \quad f_{\mathrm{body}}(\mathbf{x}_i) > \delta.
\end{split}
\end{equation}

\subsubsection{Details of Diffusion-Based Semantic-Decoupled Optimization}
Section 4.4 provides a comprehensive overview of our optimization methods, with a particular emphasis on the proposed PS-DS method. This section delves into the specifics of additional losses used in our framework. 

To optimize the skin model, we utilize the SDS method \cite{poole2022dreamfusion, stable-dreamfusion} along with a binary cross-entropy mask loss $\mathcal{L}_\mathrm{mask}$, which penalizes the difference between the silhouettes of the rendered avatar and the SMPL \cite{loper2015smpl} mesh. The total loss is formulated as:
\begin{equation}
    \mathcal{L}_\mathrm{skin} = \lambda_1 \mathcal{L}_\mathrm{SDS} + \lambda_2 \mathcal{L}_\mathrm{mask},
\end{equation}
where $\lambda_1$ and $\lambda_2$ are hyperparameters, and $\mathcal{L}_\mathrm{SDS}$ is the loss obtained from the SDS method.

For optimizing the clothing model, we employ the PS-DS method, along with an SDF loss and a pixel entropy loss. The SDF loss is used to control the clothing SDF values of each point close to its skin (body) SDF values and is formulated as:
\begin{equation}
    \mathcal{L}_\mathrm{SDF} = \mathcal{L}_\mathrm{MSE}(f_\mathrm{cloth}(\mathbf{x}_i), f_\mathrm{body}(\mathbf{x}_i)),
\end{equation}
where $\mathcal{L}_\mathrm{MSE}$ is the mean squared error loss function, $f_\mathrm{cloth}(\mathbf{x}_i)$ and $f_\mathrm{body}(\mathbf{x}_i)$ are the SDF values of the clothing and skin models, respectively. Here we fix the parameters of the skin model, making $f_\mathrm{body}(\mathbf{x}_i)$ a detached value.

To further improve our clothing model, we introduce the pixel entropy loss that considers the proportion of pixel colors from the skin and clothing models. The purpose of this loss is to ensure that one component dominates the pixel color, which allows us to control the thickness of the clothing.

To incorporate the entropy loss into our approach, we begin by computing the contributions of the skin model, denoted as $p_\mathrm{body}$, and the clothing model, denoted as $p_\mathrm{cloth}$, to each pixel color. Specifically, in Dual Volume Rendering, we update the $\alpha_i$ and $\mathbf{c}i$ parameters of the volume rendering equation $\hat{C}=\sum^{n}_{i=1} {T_{i} \alpha_{i} \mathbf{c}_{i}}$ using the merge space properties. This allows us to separate the contributions of the two models as follows:
\begin{equation}
\begin{aligned}
    T_{i} \alpha_{i} \mathbf{c}_{i} = w_{\mathrm{body},i} \mathbf{c}_\mathrm{body}(\mathbf{x}_i) + w_{\mathrm{cloth},i} \mathbf{c}_\mathrm{cloth}(\mathbf{x}_i),
\end{aligned}
\end{equation}
where $w_{\mathrm{body},i}$ and $w_{\mathrm{cloth},i}$ represent the calculated color weights. Subsequently, we can derive $p_\mathrm{body}$ and $p_\mathrm{cloth}$ as follows:
\begin{equation}
\begin{aligned}
    p_{\mathrm{body}} = \frac{\sum^{n}_{i=1} w_{\mathrm{body},i}}{\sum^{n}_{i=1} w_{\mathrm{body},i} + \sum^{n}_{i=1} w_{\mathrm{cloth},i}},
\end{aligned}
\end{equation}
\begin{equation}
\begin{aligned}
    p_{\mathrm{cloth}} = \frac{\sum^{n}_{i=1} w_{\mathrm{cloth},i}}{\sum^{n}_{i=1} w_{\mathrm{body},i} + \sum^{n}_{i=1} w_{\mathrm{cloth},i}}.
\end{aligned}
\end{equation}
After obtaining the contributions, we can compute the entropy of these proportions using the following equation:
\begin{equation}
    \mathcal{L}_\mathrm{entropy} = -p_\mathrm{body}\mathrm{log}(p_\mathrm{body}) - p_\mathrm{cloth}\mathrm{log}(p_\mathrm{cloth}).
\end{equation}

The total loss for optimizing the clothing model is given by:
\begin{equation}
    \mathcal{L}_\mathrm{clothing} = \lambda_3 \mathcal{L}_\mathrm{PS \mbox{-} DS} + \lambda_4 \mathcal{L}_\mathrm{SDF} + \lambda_5 \mathcal{L}_\mathrm{entropy},
\end{equation}
where $\lambda_3$, $\lambda_4$, and $\lambda_5$ are hyperparameters, and $\mathcal{L}_\mathrm{PS \mbox{-} DS}$ is the loss obtained from the PS-DS method.

\subsection{Implementation Details}
For volume rendering during training, we sample 32 points on each ray and set the scaling factor in $\Phi_s$, to a fixed value of $s=e^7$. This fixed value ensures that the residual SDF value does not dominate the overall proportion.
For the skin model optimization, we set the hyperparameters $\lambda_1 = 100$ and $\lambda_2 = 800$ to balance the contributions of the SDS loss and the mask loss. We optimize the skin model for a total of 25000 iterations. To update the model parameters, we use a learning rate of $1e-4$.
Subsequently, we focus on optimizing the clothing model. We set the hyperparameters $\lambda_3 = 100$, $\lambda_4 = 300$, $\lambda_5 = 0$, and $\delta = 1e-4$. This optimization process for the clothing model also runs for 25000 iterations. The learning rate for this stage is set to $5e-4$.
To further refine the generated clothing and filter out any noise, we perform an additional optimization stage. We set the hyperparameters as $\lambda_3 = 100$, $\lambda_4 = 300$, $\lambda_5 = 5000$, and $\delta = 2e-3$. The purpose of this stage is to facilitate the separation of clothing from body. The learning rate for this stage is set to $5e-4$. Similar to the previous stages, we conduct this optimization for 25000 iterations.
Our model training is carried out on an NVIDIA Tesla V100 32GB GPU.

\subsection{More Results}
We present additional comparison results with baselines in Figure \ref{fig:baseline_more}, detailed results of AvatarFusion in Figure \ref{fig:results_more}. Due to page limitations, we have provided additional results on the project page and an internet image repository. We present more ablation results on \href{https://postimg.cc/vDxRTnpj}{https://postimg.cc/vDxRTnpj} and \href{https://postimg.cc/zyB5wFpk}{https://postimg.cc/zyB5wFpk}. For further comparison, we include results with Stable Diffusion 2D on \href{https://postimg.cc/rRqpKXWj}{https://postimg.cc/rRqpKXWj} and backside comparison results on \href{https://postimg.cc/c6HyVFSq}{https://postimg.cc/c6HyVFSq}.






\end{document}